\documentclass[10pt,twocolumn,letterpaper]{article}

\usepackage[pagenumbers]{cvpr} 

\newcommand{\D}{\mathcal{D}}

\newcommand{\G}{\mathcal{G}}

\newcommand{\simc}{\operatorname{sim}}
\newcommand{\norm}[1]{\left\lVert#1\right\rVert_2}

\usepackage{multirow}
\usepackage{makecell}
\usepackage{xcolor}
\usepackage{colortbl}
\usepackage{tikz}
\usepackage{amsmath}
\usepackage{amssymb}

\definecolor{falseNeg}{HTML}{3d9af5}
\definecolor{falsePos}{HTML}{fa5757}

\definecolor{gold}{HTML}{FBF2D2}
\definecolor{silver}{HTML}{DDDDDD}
\definecolor{bronze}{HTML}{EED2B8}

\definecolor{goldD}{HTML}{D9AE13}
\definecolor{silverD}{HTML}{909090}
\definecolor{bronzeD}{HTML}{9A5F26}

\newcommand{\medal}[3]{\tikz[baseline=(char.base)]{\node[rounded corners=2pt,fill=#1,draw=#2,inner sep=1pt](char){#3};}}

\newcommand{\bm}[2]{
    \ifcase#1\or
      {\medal{gold}{goldD}{\textbf{#2}}}
    \or 
      {\medal{silver}{silverD}{#2}}
    \or 
      {#2}
    \else 
      #2
    \fi\ignorespaces
}

\definecolor{weights}{HTML}{0398fc}
\colorlet{vissamp}{blue!8}

\newcommand{\dpos}[1]{\textcolor{green!50!black}{+#1}}
\newcommand{\dneg}[1]{\textcolor{red!70!black}{$-$#1}}

\usepackage{tabularx}

\definecolor{cvprblue}{rgb}{0.21,0.49,0.74}
\usepackage[pagebackref,breaklinks,colorlinks,allcolors=cvprblue]{hyperref}

\title{One Representation to Rule Them All -- Open-Vocabulary Change Detection from Text or Visual Examples}

\author{Blaž Rolih
\and Matic Fučka
\and Filip Wolf
\and Luka Čehovin Zajc
\and University of Ljubljana, Faculty of Computer and Information Science, Slovenia\\
{\tt\small blaz.rolih@fri.uni-lj.si}
}

\begin{document}
\maketitle
\begin{abstract}
Open-vocabulary change detection (OVCD) aims to localise user-specified changes without retraining for predefined categories. Existing approaches typically rely on textual descriptions and combine separate models for region proposal, semantic recognition, and temporal comparison. This limits how target concepts can be expressed and makes detection dependent on several intermediate predictions. We introduce OVTSID3, a training-free framework that unifies region grouping, concept identification, and temporal comparison in the representation space of DINOv3. OVTSID3 supports two complementary modes of concept specification: zero-shot textual queries or one-shot visual prompting through a masked reference pair. OVTSID3 outperforms all prior training-free OVCD methods on two building and two natural-disaster benchmarks, raising the average F1 by 8.9 points. OVTSID3 also supports semantic change detection and achieves state-of-the-art on SECOND. These results demonstrate the flexibility of a unified representation for text- or example-guided change detection without task-specific model training. Code: \textcolor{magenta}{TBA}

\end{abstract}    
\section{Introduction}
\label{sec:intro}

Change detection (CD) is a key task in remote sensing (RS), with applications ranging from urban expansion monitoring to natural disaster response~\cite{rolih2026mason, daudt2018fcn, ding2025labelEffsurvey}. In a typical setting, two images of the same geographical region acquired at different times are compared to localise a \textit{specific type of change}~\cite{rolih2026changeflow, cheng2025changedino, chen2024changeMamba}. Conventional supervised approaches define change types through their training data and therefore operate in a \textit{closed-vocabulary} setting. Adapting a detector to a new concept, \eg, moving from building to water change detection, requires additional annotations and retraining. This dependence is particularly restrictive in applications such as natural disaster response, where labelled examples can be difficult to acquire~\cite{rolih2026mason, valsamis2024s2wcd}. It also limits the use of such detectors as reusable perception tools in agentic systems, where the subject of analysis may change with the user's request~\cite{rasheed2026hard}.

\begin{figure}
    \centering
    \includegraphics[width=1\linewidth]{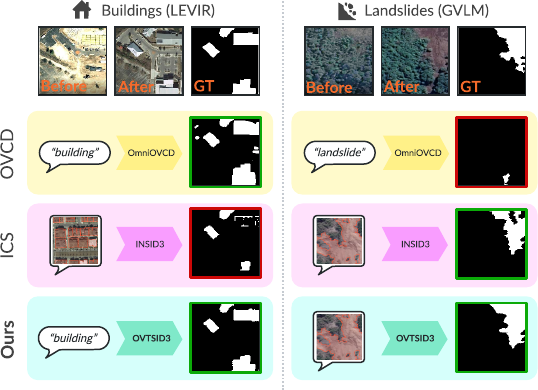}
    \caption{\textbf{Textual and visual concept specification for change detection.} Text-only OVCD methods can detect common concepts such as buildings but struggle with Earth Observation concepts whose appearance is poorly captured by textual prompts, such as landslides. \textit{Visual in-context segmentation} (ICS) offers an alternative way to specify these targets, but is limited to single images.
    OVTSID3 supports change detection through both textual descriptions or masked visual exemplars to achieve good performance in both scenarios.}
    \label{fig:title}
\end{figure}

Recent work instead adopts \textit{open-vocabulary change detection} (OVCD), allowing users to specify target concepts at inference time~\cite{guo2026opendpr, li2025dynearth, su2026seg2change}. Some methods chain together foundation models such as SAM~\cite{kirillov2023sam}, DINO~\cite{oquab2024dinov2}, and SegEarth-OV~\cite{li2025segearth} for mask proposal, temporal comparison, and concept identification~\cite{li2025dynearth, guo2026opendpr, su2026seg2change, dou2026adaptovcd}. This allows errors in any stage of the chain to propagate to the final prediction. Other models, instead, unify all steps within a single SAM3~\cite{carion2025sam3} architecture~\cite{zhang2026omniovcd} but encounter two failure cases that are particularly relevant to Earth Observation (EO).

Firstly, SAM3 achieves suboptimal performance in segmenting EO concepts (see \cref{fig:qual}) because it was not designed for this domain, resulting in errors such as fragmented masks or over-segmented regions with unclear boundaries. Here, subsequent filtering cannot recover omitted pixels and can even discard correct regions, since some of them span irrelevant concepts due to over-segmentation.

\begin{figure}[t]
    \centering
    \includegraphics[width=1\linewidth]{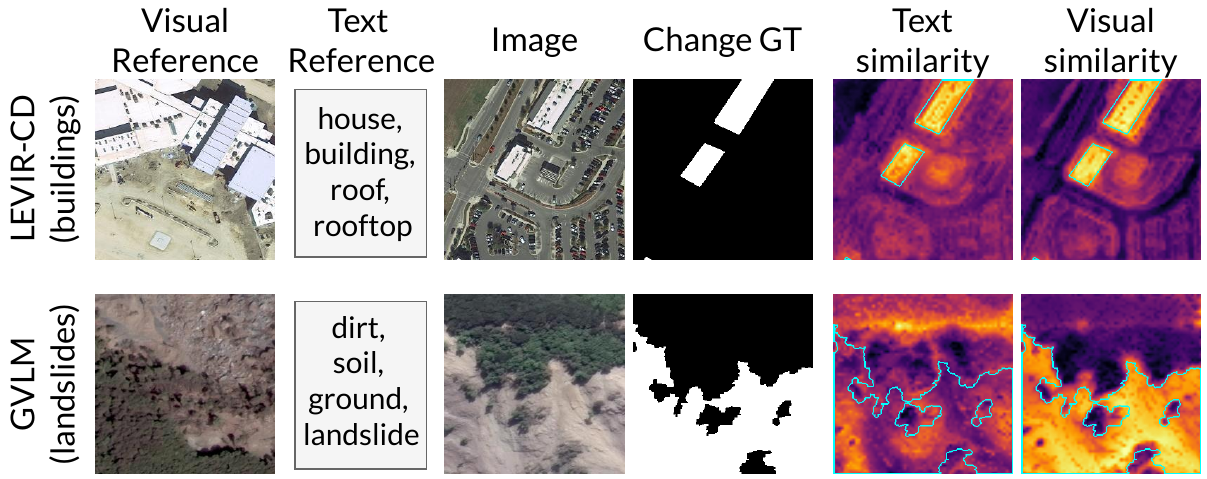}
    \caption{While textual prompts respond well to common objects that are easy to refer to, like a "building", they provide weak responses for EO concepts that are not directly in the vocabulary and not easy to describe, such as "landslide". Using a prompt based on visual concept reference (explained in \Cref{par:vis_prompt}) solves this, as the query comes from the same encoder and produces responses even for concepts not in the text vocabulary.}
    \label{fig:corr}
\end{figure}

Secondly, even correctly segmented regions can be misidentified if they exhibit poor image-text alignment (\eg, landslides in \cref{fig:corr}). A natural alternative for concepts that are difficult to specify through text is to provide a \textit{visual example}. While some existing CD works explore this~\cite{guo2026opendpr}, they still generate visual prototypes from textual descriptions and retain a SAM-based segmentation pipeline, leaving both issues unaddressed. \textit{Visual in-context segmentation} instead offers a more direct approach. INSID3~\cite{cuttano2026insid3}, for example, segments an image using a reference pair of similar images and a mask specifying the concept of interest. While promising, INSID3 addresses only single-image segmentation and requires an annotated visual reference, even when a textual description would suffice. This motivates a unified change-detection framework that retains reference-free textual prompting while supporting visual examples for concepts that are poorly captured by language.

To address this, we propose \textbf{OVTSID3} (\textbf{O}pen-\textbf{V}ocabulary \textbf{T}ext- or \textbf{S}ample-\textbf{I}mage-prompted change \textbf{D}etection with DINOv\textbf{3}), which is built on the representations of a single frozen DINOv3 backbone~\cite{simeoni2025dinov3} by using them for concept region grouping, concept identification, and change estimation. By fully exploiting the DINOv3 latent space, the approach requires no model chaining and no reliance on mask proposal networks (\eg, SAM). Additionally, OVTSID3 supports both textual and visual prompting as illustrated in \cref{fig:title}. For textual prompting, we use the official text encoder paired with the visual encoder~\cite{jose2025dinotxt}, while for visual prompting, we construct a concept prototype directly from the features of a single masked reference pair, allowing the target to be specified by its appearance. Both modalities produce concept scores over the same visual features and feed into a shared cluster-based selection procedure. Feature differences provide complementary, concept-agnostic evidence of change, which is combined with the selected concept regions to produce the final prediction. Importantly, this formulation is training-free and requires neither a separate mask-proposal network nor task-specific model training.

OVTSID3 achieves state-of-the-art results and outperforms the previous best training-free open-vocabulary method on the building-change benchmarks LEVIR-CD~\cite{chen2020levirStanet} and WHU~\cite{ji2018whu}, as well as on landslide changes in GVLM~\cite{zhang2023gvlm} and wildfire-related changes in S2WCD~\cite{valsamis2024s2wcd}. It also outperforms the previous best method on semantic CD, evaluated on SECOND~\cite{yang2020second}. These results demonstrate the effectiveness of a shared representation across both text-guided detection of common concepts and visually guided detection of natural-disaster changes.

Our contributions are threefold:
\begin{itemize}
    \item We extend the open-vocabulary change detection paradigm with visual prompting, enabling strong performance in scenarios where textual descriptions do not reliably capture a concept's appearance.
    \item We unify temporal change comparison with textual and visual change filtering within the same DINOv3 latent space.
    \item We demonstrate strong zero-shot text-guided change detection across buildings and one-shot visually guided change detection across natural-disaster benchmarks without task-specific model training.
\end{itemize}

\section{Related work}
\label{sec:rel}

\noindent\textbf{Remote Sensing Change Detection.}
Remote sensing change detection (CD) has evolved from pixel-wise differencing and statistical techniques~\cite{singh1989reviewCD,bovolo2006theoretical} to deep learning approaches based on Siamese convolutional networks~\cite{daudt2018fcn,chen2020levirStanet}, transformers~\cite{chen2021bit,bandara2022changeFormer,yu2024maskcd, rolih2025btc}, and state-space models~\cite{chen2024changeMamba}. Despite their effectiveness, most deep CD methods rely on supervised training with annotated bi-temporal images, limiting their applicability to unseen environments and change categories~\cite{ruuvzivcka2022ravaen,hertel2025rapid}. Unsupervised CD (UCD) alleviates this dependency through feature comparison~\cite{saha2019unsupervised,du2019sfa}, synthetic changes~\cite{chen2023_i3pe,song2024syntheworld,rolih2026mason}, or pretrained visual foundation models~\cite{oquab2024dinov2, simeoni2025dinov3, kirillov2023sam, wolf2026deo}. In particular, AnyChange~\cite{zheng2024anychange} exploits the segmentation capabilities of SAM~\cite{kirillov2023sam} in a training-free manner. However, conventional CD and UCD generally produce change masks with predefined types or are type-agnostic, thereby preventing users from specifying which semantic categories are of interest.

\noindent\textbf{Open-Vocabulary Change Detection.}
Open-vocabulary change detection (OVCD) extends CD beyond a fixed set of types by allowing changes to be identified according to user-specified ones. Early approaches, such as SCM~\cite{tan2024scm}, combine segmentation foundation models with vision-language representations for type-specific change identification. DynamicEarth~\cite{li2025dynearth} formalises OVCD and introduces two complementary training-free paradigms: discovering changes before identifying their categories, or first identifying semantic regions and subsequently detecting their changes. Its implementations combine multiple foundation models for segmentation, feature comparison, and semantic recognition.

OpenDPR~\cite{guo2026opendpr} identifies semantic classification as a major bottleneck of such pipelines, particularly for fine-grained land-cover categories that are poorly represented by vision-language models. To address this limitation, it constructs diverse visual prototypes using diffusion-generated imagery and retrieves matching change proposals directly in the visual feature space. While reducing reliance on text-image similarity during inference, its prototype generation remains conditioned on textual category descriptions. 

\begin{figure*}[!t]
    \centering
    \includegraphics[width=1\linewidth]{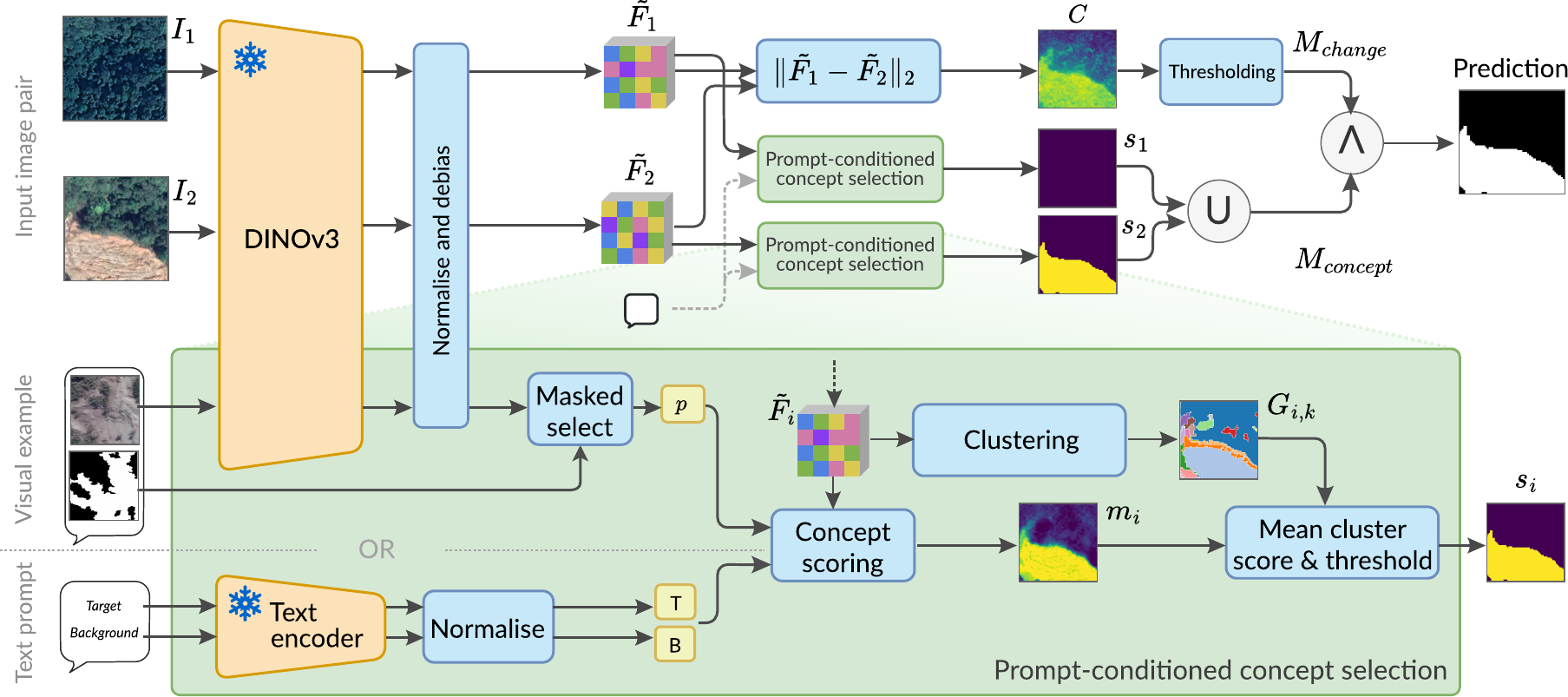}
    \caption{\textbf{Overview of OVTSID3}. Both images are encoded with a frozen DINOv3 backbone, and their features are positionally debiased (\Cref{sec:features}). Each feature map is grouped into clusters $\G_i$ (\Cref{sec:clustering}), and every location receives a concept score $m_i$ from either a text prompt or a masked visual reference (\Cref{sec:prompts}). Clusters whose mean score $a_{i,k}$ exceeds a threshold form the concept masks $S_1$ and $S_2$ (\Cref{sec:concept_mask}). Independently of the prompt, the feature difference gives a class-agnostic change map $C$ (\Cref{sec:changedetection}), and the final prediction keeps changed locations that contain the concept in either image (\Cref{sec:postproc}).}
    
    \label{fig:arch}
\end{figure*}

Recent work further explores DINOv3-based change estimation~\cite{dou2026adaptovcd} and reusable class-agnostic change evidence~\cite{wang2026cogvis}. OmniOVCD~\cite{zhang2026omniovcd} removes the model chaining and performs everything within a single SAM3~\cite{carion2025sam3}-based model. While this addresses the failure point of model chaining, it still relies on SAM for segmentation, which fails on certain EO concepts such as landslide-affected areas. It also uses only text-based filtering, which fails when the correspondence between text and concepts is poor. 

In contrast, our approach performs concept identification and change localisation directly from a shared visual representation. It supports both textual descriptions and user-provided visual exemplars without requiring synthetic prototype generation or a separate change-detection network.

\section{OVTSID3}

Given co-registered images captured at different times and a user-specified target prompt, such as text descriptions of foreground and background or a masked visual reference, the model must produce a change mask corresponding to that concept. For this, our framework uses a single frozen DINOv3~\cite{simeoni2025dinov3} backbone for \textbf{(i)} grouping locations with similar features, \textbf{(ii)} concept matching, and \textbf{(iii)} estimating temporal changes. 
Regardless of the prompt modality, both paths use the same cluster-based concept selection and concept-agnostic temporal change map, as explained in the following sections and shown in \Cref{fig:arch}.

\subsection{Dense Visual Features and Positional Debiasing}
\label{sec:features}

We extract features independently from a pair of images $(I_1, I_2)$ using the frozen DINOv3 backbone $\D$:
\begin{equation}
F_i=\D(I_i)\in\mathbb{R}^{h\times w\times d},\qquad i\in\{1,2\},
\end{equation}
followed by $\ell_2$-normalisation of both feature vectors. To remove the positional bias present in DINOv3 features, following~\cite{cuttano2026insid3}, we construct a basis from the feature map of an all-zero input image obtained with $\D$. Let $E\in\mathbb{R}^{d\times hw}$ contain these normalised features with the mean over spatial positions removed, and let $U_r\in\mathbb{R}^{d\times r}$ contain the first $r$ leftmost singular vectors of $E$. We construct the projection matrix and project features $F_i$ to an orthogonal complement without the positional bias:
\begin{equation}
P_{\perp}=\textbf{1}_d-U_r U_r^\top \enspace ,
\qquad
\widetilde F_i=F_iP_{\perp},
\label{eq:debias}
\end{equation}
where $\textbf{1}_d$ is an identity matrix of size $d$. Unlike previous work~\cite{cuttano2026insid3}, which uses non-debiased features for inter-image operations, we always use debiased features $\widetilde F_i$ in all further steps. This stems from the fact that remote sensing has no canonical orientation~\cite{rolf2024satellite}, so the locality induced by this bias is undesirable (\ie, the features of a house in the top-left corner should not differ from those in the bottom-right corner). We adopt the same $r=500$ as~\cite{cuttano2026insid3}.

\subsection{Clustering}
\label{sec:clustering}

For each temporal step, we separately partition the debiased feature map into clusters using agglomerative clustering~\cite{mullner2011modern} with average linkage. Specifically, the pairwise distance between feature locations $l_1$ and $l_2$ is given by
\begin{equation}
d_i(l_1,l_2)=1-\simc\bigl(\widetilde F_i(l_1),\widetilde F_i(l_2)\bigr),
\end{equation}
with merges stopping at the distance threshold $1-\tau$, where $\tau$ controls cluster granularity.

Let $\G_i=\{G_{i,1},\ldots,G_{i,K_i}\}$ be the resulting partition. Each cluster is represented by a unit-normalised mean feature
\begin{align}
c_{i,k}&=\frac{1}{|G_{i,k}|}\sum_{x\in G_{i,k}}\widetilde F_i(l).
\end{align}

\subsection{Prompt-Conditioned Concept Scores}
\label{sec:prompts}

For each image, we construct a scalar concept score $m_i(l)$ at every feature-map location $l = \{(x,y) \in H \times W\}$, with the scoring rule determined by the prompt type.

\noindent\textbf{Text prompts.}
Let $T$ and $B$ be sets of target and background descriptions encoded by the text encoder paired with the text-aligned DINOv3 model~\cite{jose2025dinotxt}. With cosine similarity $\simc(\cdot,\cdot)$, the score is realised as a foreground--background margin:
\begin{equation}
m_i^{\mathrm{txt}}(l)=
\max_{t\in T}\simc\bigl(\widetilde F_i(l),t\bigr)
-\max_{b\in B}\simc\bigl(\widetilde F_i(l),b\bigr).
\label{eq:textmargin}
\end{equation}
This allows us to measure the presence of a target relative to the background classes, rather than relying solely on target features. This directly enables text-based zero-shot usage, which INSID3~\cite{cuttano2026insid3} does not support.

\noindent\textbf{Visual prompts.}
\label{par:vis_prompt}
We can also prompt the model with a reference pair of the concept together with its binary change mask (\eg, an annotated landslide) when concepts are hard to describe or have weak text correspondence, as shown in \Cref{fig:corr}.
    
Let $\widetilde F^r$ denote the normalised, debiased features of the post-change reference image, $M^r$ the reference mask resized to feature resolution, $\Omega^r=\{l: M^r(l)=1\}$ the masked locations, and $c^r$ their normalised mean feature described in \Cref{sec:clustering}. The prototype $p$ is the mean of the fraction $q$ of masked locations most similar to this centroid $c^r$, which form the core of the masked concept:
\begin{equation}
\begin{aligned}
\Omega^r_q &= \operatorname*{arg\,top\text{-}k}_{l\in\Omega^r}\ \simc\bigl(\widetilde F^r(l), c^r\bigr), \quad k=\lceil q|\Omega^r|\rceil,\\
p &= \frac{1}{|\Omega^r_q|}\sum_{l\in\Omega^r_q}\widetilde F^r(l).
\end{aligned}
\end{equation}
Taking only a fraction $q$ of the masked location $\Omega^r$ most similar to the centroid $c^r$ makes the prototype $p$ contain less noise, as certain masks of concepts contain features unrelated to the desired concept (\eg, a mask of a landslide might contain a grass patch). 
This also has the added effect of making the prototype insensitive to small perturbations in features.
We use $q=0.25$, with results on the GVLM validation set being stable for $q\in[0.05,0.5]$ (see Supplementary). Unlike with textual prompts, we do not treat the region outside the mask as the background, since this would result in a single background type.

In the case of visual prompts, the concept score $m_i$ is the similarity of each location with the prototype $p$. Since similarities depend on the scene and acquisition conditions of each image, we centre them by an image-level offset. To do this, we first compute an image-level mean similarity $\bar s_i=\frac{1}{hw}\sum_{x}\simc\bigl(\widetilde F_i(l),p\bigr)$ to the prototype for both time-steps. The minimum of two estimates the response of non-concept content, and we subtract it from every similarity:
\begin{equation}
m_i^{\mathrm{vis}}(l)=\simc\bigl(\widetilde F_i(l),p\bigr)-\min\bigl(\bar s_1,\bar s_2\bigr).
\label{eq:visualsingle}
\end{equation}
Taking the lower mean prevents the concept from being centred away when it covers most of one image (\eg, a large burn scar), and a single offset per pair keeps both images on the same scale for the shared threshold $\delta$.

The two prompt modalities share subsequent processing, so we use $m_i$ to refer to either score. 

\subsection{Concept Mask Selection}
\label{sec:concept_mask}

Each cluster is scored by the mean concept score from \Cref{sec:prompts} over its locations:
\begin{equation}
a_{i,k}=\frac{1}{|G_{i,k}|}\sum_{l\in G_{i,k}}m_i(l).
\label{eq:cluster_score}
\end{equation}
With $g_i(l)$ denoting the cluster assignment at location $l$, the concept-presence mask keeps every cluster whose score exceeds a threshold $\delta$:
\begin{equation}
S_i(l)=\bigl[a_{i,g_i(l)}>\delta\bigr].
\label{eq:conceptmask}
\end{equation}
Deciding per cluster rather than per location ensures that all locations within a homogeneous region share the decision, which suppresses isolated noisy responses and recovers parts of an object whose individual scores are weak.
For visual prompts, $\delta$ is calibrated on the reference pair: we score and cluster the reference pair as stated above and set $\delta$ to the mean cluster score outside $M^r$, \ie, the typical response of non-concept regions. For text prompts, the margin is already centred at zero, and $\delta$ is a small constant (\Cref{sec:impl}).

Since the concept can exist for either image $I_i$ (\eg appearance, disappearance), we use the union of both to form the final concept mask:
\begin{equation}
M_{\mathrm{concept}}=S_1\cup S_2.
\label{eq:conceptunion}
\end{equation}

\subsection{Class-Agnostic Change Estimation}
\label{sec:changedetection}

Independently of the prompt type, we compute the magnitude of the difference between corresponding debiased features:
\begin{equation}
C(l)=\norm{\widetilde F_1(l)-\widetilde F_2(l)}.
\label{eq:cmap}
\end{equation}
The class-agnostic change mask is then given by 
\begin{equation}
M_{\mathrm{change}}(l)=C(l)>\theta,
\label{eq:changebinary}
\end{equation}
where $\theta$ is the threshold produced by Otsu's method~\cite{otsu1979threshold}.

\subsection{Mask Fusion and Post-processing}
\label{sec:postproc}

We intersect the change and concept masks to obtain the final change mask:
\begin{equation}
M(l)=M_{\mathrm{change}}(l)\land M_{\mathrm{concept}}(l).
\label{eq:final}
\end{equation}
This combination is more robust than simply performing open-vocabulary segmentation because the type-agnostic change mask does not depend on the success of the prompt alone. It will always produce a mask based solely on whether features at that location differ. This leads to better delineation of changed regions, where the border cannot be easily separated by a prompt alone.

\subsection{Extension to Semantic Change Detection}
\label{sec:scd}

The same components extend directly to semantic change detection, where each changed location must additionally be assigned its land-cover class in both images. Instead of a single target concept, the user specifies $C$ classes, each given by a set of text descriptions $T_c$, $c\in\{1,\ldots,C\}$. Each description is embedded using the text encoder.

\noindent\textbf{Class scores.}
At every location, the score of class $c$ is the mean similarity to its $k$ best-matching descriptions:
\begin{equation}
s_{i,c}(l)=\frac{1}{k}\max_{\substack{T'\subseteq T_c\\|T'|=k}}\;\sum_{t\in T'}\simc\bigl(\widetilde F_i(l),t\bigr).
\label{eq:clsscore}
\end{equation}
Averaging over the top $k$ (in our case top 3) descriptions rather than taking the maximum prevents classes with more descriptions from being favoured, since the maximum over a larger set is biased upwards.

\noindent\textbf{Cluster-level classification.}
Class assignment is performed on the clusters from Section~\ref{sec:clustering}, which makes all locations of a homogeneous region share a label. Each cluster takes the class with the highest mean score, and each location inherits the class of its cluster:
\begin{equation}
\hat y_{i,k}=\arg\max_{c}\frac{1}{|G_{i,k}|}\sum_{l\in G_{i,k}}s_{i,c}(l),
\qquad
y_i(l)=\hat y_{i,g_i(l)}.
\label{eq:clsassign}
\end{equation}
Since every location receives one of the user-specified classes, the concept mask selection described in Section~\ref{sec:concept_mask} is unnecessary.

\noindent\textbf{Semantic change mask.}
A location is semantically changed if its features change and its class differs between the two images:
\begin{equation}
M_{\mathrm{sem}}(l)=M_{\mathrm{change}}(l)\land\bigl[y_1(l)\neq y_2(l)\bigr].
\label{eq:semgate}
\end{equation}
The class-agnostic map $M_{\mathrm{change}}$ captures where the scene changed, while the class condition discards changes that do not alter the land cover, such as illumination, season, or acquisition differences. The final semantic maps for each time-step assign $y_i(l)$ to changed locations (i.e., $M_{\mathrm{sem}}(l) = 1$) and the no-change label to all others.

\definecolor{oursblue}{HTML}{E6EDF7}
\begin{table*}[!t]
    \centering
        \caption{\textbf{Change detection results on building changes (LEVIR-CD and WHU), landslides (GVLM), and wildfire changes (S2WCD).} Prompt: a text prompt or a single masked visual reference (same references used for all visual-prompt methods). \textit{Avg} denotes the mean over all four datasets. The \colorbox{gold}{best} and \colorbox{silver}{second-best} results are highlighted. Standard deviation is reported in the Supplementary where applicable. $*$ indicates the method was trained on in-domain data.}
    
    \label{tab:sota_ovcd}
    \setlength{\tabcolsep}{4pt}
    \begin{tabular}{lccccccccccc}
    \toprule
    	\multirow{2}{*}{~} & \multirow{2}{*}{Prompt} & \multicolumn{2}{c}{LEVIR} & \multicolumn{2}{c}{WHU} & \multicolumn{2}{c}{GVLM} & \multicolumn{2}{c}{S2WCD} & \multicolumn{2}{c}{\textit{Avg}}\\
~ & ~ & IoU & F1 & IoU & F1 & IoU & F1 & IoU & F1 & IoU & F1\\
 \midrule
\multicolumn{12}{l}{\textit{Unsupervised}}\\
Pixel Difference & -- & 4.7 & 8.9 & 7.9 & 14.7 & 13.3 & 23.4 & 24.1 & 38.8 & 12.5 & 21.4\\
S2C (DINOv3)$^*$~\cite{ding2025s2c}\tiny{AAAI26} & -- & 20.2 & 33.7 & 17.3 & 29.4 & 27.1 & 42.7 & 0.0 & 0.0 & 16.2 & 26.4\\
AnyChange~\cite{zheng2024anychange}\tiny{NeurIPS24} & -- & 14.6 & 25.4 & 12.4 & 22.0 & 11.6 & 20.8 & 4.2 & 8.0 & 10.7 & 19.1\\
\quad$\hookrightarrow$ point query & Point & 15.6 & 27.0 & 23.2 & 37.6 & 6.7 & 12.5 & 0.7 & 1.4 & 11.6 & 19.6\\
Mason$^*$~\cite{rolih2026mason}\tiny{TGRS26} & -- & 24.8 & 39.7 & 20.3 & 33.8 & \bm1{38.3} & \bm1{55.4} & 33.8 & 50.3 & 29.3 & 44.8\\
 \midrule
\multicolumn{12}{l}{\textit{Supervised open-vocabulary}}\\
\color{gray}Seg2Change~\cite{su2026seg2change}\tiny{ACMM26} & \color{gray}Text & \color{gray}64.9 & \color{gray}78.7 & \color{gray}75.7 & \color{gray}86.2 & \color{gray}44.1 & \color{gray}61.2 & \color{gray}2.4 & \color{gray}4.7 & \color{gray}46.8 & \color{gray}57.7\\
 \midrule
\multicolumn{12}{l}{\textit{In-context segmentation and mask XOR}}\\
INSID3~\cite{cuttano2026insid3}\tiny{CVPR26} & Visual & 27.5 & 43.2 & 36.1 & 53.0 & 22.6 & 36.7 & 25.1 & 40.0 & 27.8 & 43.2\\
\quad$\hookrightarrow$ + our change map & Visual & 45.3 & 62.2 & 52.0 & 68.3 & 38.0 & 55.1 & 55.0 & 70.2 & 47.6 & 64.0\\
 \midrule
\multicolumn{12}{l}{\textit{Training-free open-vocabulary}}\\
SCM~\cite{tan2024scm}\tiny{IGARSS24} & Text & 15.8 & 27.3 & 20.0 & 33.3 & 16.3 & 28.0 & 5.4 & 10.2 & 14.4 & 24.7\\
DynEarth M-C-I~\cite{li2025dynearth}\tiny{AAAI26} & Text & 32.9 & 49.5 & 57.3 & 72.9 & 16.4 & 28.2 & 6.0 & 11.3 & 28.1 & 40.4\\
DynEarth I-M-C~\cite{li2025dynearth}\tiny{AAAI26} & Text & 20.0 & 33.3 & 8.7 & 16.0 & 0.7 & 1.3 & 1.0 & 2.0 & 7.6 & 13.2\\
OpenDPR~\cite{guo2026opendpr}\tiny{CVPR26} & Text & 44.8 & 61.9 & 55.6 & 71.4 & 21.6 & 35.5 & 10.2 & 18.5 & 33.0 & 46.9\\
OmniOVCD~\cite{zhang2026omniovcd}\tiny{arXiv26} & Text & \bm2{67.8} & \bm2{80.8} & \bm2{65.7} & \bm2{79.3} & 21.9 & 35.9 & 32.0 & 48.5 & 46.9 & 61.1\\
\rowcolor{oursblue}
\textbf{OVTSID3} (visual) & Visual & 41.9 & 58.4 & 61.0 & 75.5 & \bm2{38.2} & \bm2{55.3} & \bm1{58.0} & \bm1{73.4} & \bm2{49.8} & \bm2{65.6}\\
\rowcolor{oursblue}
\textbf{OVTSID3} (text) & Text & \bm1{71.0} & \bm1{83.0} & \bm1{67.7} & \bm1{80.8} & 28.3 & 44.1 & \bm2{56.2} & \bm2{71.9} & \bm1{55.8} & \bm1{70.0}\\

    \bottomrule
    \end{tabular}

\end{table*}

\section{Experiments}

\noindent\textbf{Datasets and evaluation.} We evaluate binary change detection on LEVIR-CD~\cite{chen2020levirStanet} and WHU~\cite{ji2018whu} for buildings, GVLM~\cite{zhang2023gvlm} for landslides, and S2WCD~\cite{valsamis2024s2wcd} for wildfire-related changes. Following recent work~\cite{guo2026opendpr, li2025dynearth, zhang2026omniovcd}, we report foreground intersection over union (IoU) and F1, together with the arithmetic mean across all four datasets. We evaluate SCD on SECOND~\cite{yang2020second} and report class-wise IoU and F1, along with means across classes~\cite{guo2026opendpr, zhang2026omniovcd}. All scores are percentages; differences are given in percentage points. For each dataset, all results are obtained on the entire public test split if available. For S2WCD with no official split, we divide it in half for train/test and ensure no event overlap (see Supplementary for more details).

\noindent\textbf{Prompting protocol.} Each visual prompt consists of one reference pair and its binary concept mask, sampled from training examples with $40$--$60\%$ foreground coverage. We construct the prototype as described in \Cref{sec:prompts}. Visual-prompt experiments are repeated with 3 reference-sampling seeds, using the same reference for all visual methods within each run, and are reported as mean with standard deviation in the Supplementary. Thus, visual prompting uses one annotated example without model fine-tuning, whereas text prompting requires no annotated reference. SECOND uses text-defined classes, which are predicted jointly using argmax. More details are in the Supplementary.

\subsection{Implementation details}
\label{sec:impl}

We use a frozen DINOv3~\cite{simeoni2025dinov3} ViT-Large with text alignment as in~\cite{jose2025dinotxt}. Inputs are resized to $1024\times1024$, following INSID3~\cite{cuttano2026insid3} and OmniOVCD~\cite{zhang2026omniovcd}. Predictions are bilinearly resized to the original resolution and refined using CRF~\cite{krahenbuhl2011crf} as in INSID3~\cite{cuttano2026insid3}. We use agglomerative clustering with $\tau=0.8$ and a text threshold of $\delta=0.01$. These were obtained on the validation sets of LEVIR-CD and GVLM and are kept constant for all datasets (see Supplementary). The visual threshold $\delta$ is automatically estimated from the reference pair (\Cref{sec:concept_mask}).

\subsection{Results}

\noindent\textbf{Binary change detection.}
\Cref{tab:sota_ovcd} compares OVTSID3 with related methods from various paradigms. OVTSID3 uniquely supports both reference-free textual and one-shot visual prompting among the evaluated methods. Text performs better for well-defined concepts such as buildings, whereas visual references improve detection of difficult-to-describe concepts such as landslides.

\noindent\textbf{Text-prompted results.}
OVTSID3 with a textual prompt reaches 83.0 F1 on LEVIR-CD and 80.8 on WHU, the best among training-free methods and 2.2 and 1.5 points above OmniOVCD with the same prompt text. Averaged over all four datasets, it improves over OmniOVCD by 8.9 points. On GVLM, the 8.2-point gain with an identical prompt reflects SAM3's poor transfer to EO concepts and the benefit of our mechanism, which directly produces masks from clustered DINOv3 features. A visual reference in OVTSID3 adds a further 11.2 points, as landslides exhibit a poor text-image correlation (\Cref{fig:corr}).

\noindent\textbf{Visually-prompted results.}
With a single masked reference (mean over three seeds), OVTSID3 achieves 55.3 F1 on GVLM and 73.4 on S2WCD, outperforming base INSID3 (36.7 and 40.0 F1, respectively). This demonstrates the importance of explicit temporal change estimation over simply comparing segmentation masks. To isolate this contribution, we combine INSID3 with our class-agnostic change map, increasing its average F1 to 64.0, compared with 65.6 for our visual version. This indicates that temporal change evidence accounts for much of the improvement, while our visual concept selection provides additional gains. On LEVIR-CD, INSID3 with our change map outperforms our visual branch, but remains substantially below text-prompted OVTSID3 (62.2 vs.\ 83.0 F1). These results highlight the complementary nature of the two modalities: textual prompts enable effective detection of well-defined concepts, while visual prompts provide an alternative when text-image correspondence is unreliable.

\begin{figure}[t]
    \centering
    \includegraphics[width=1\linewidth]{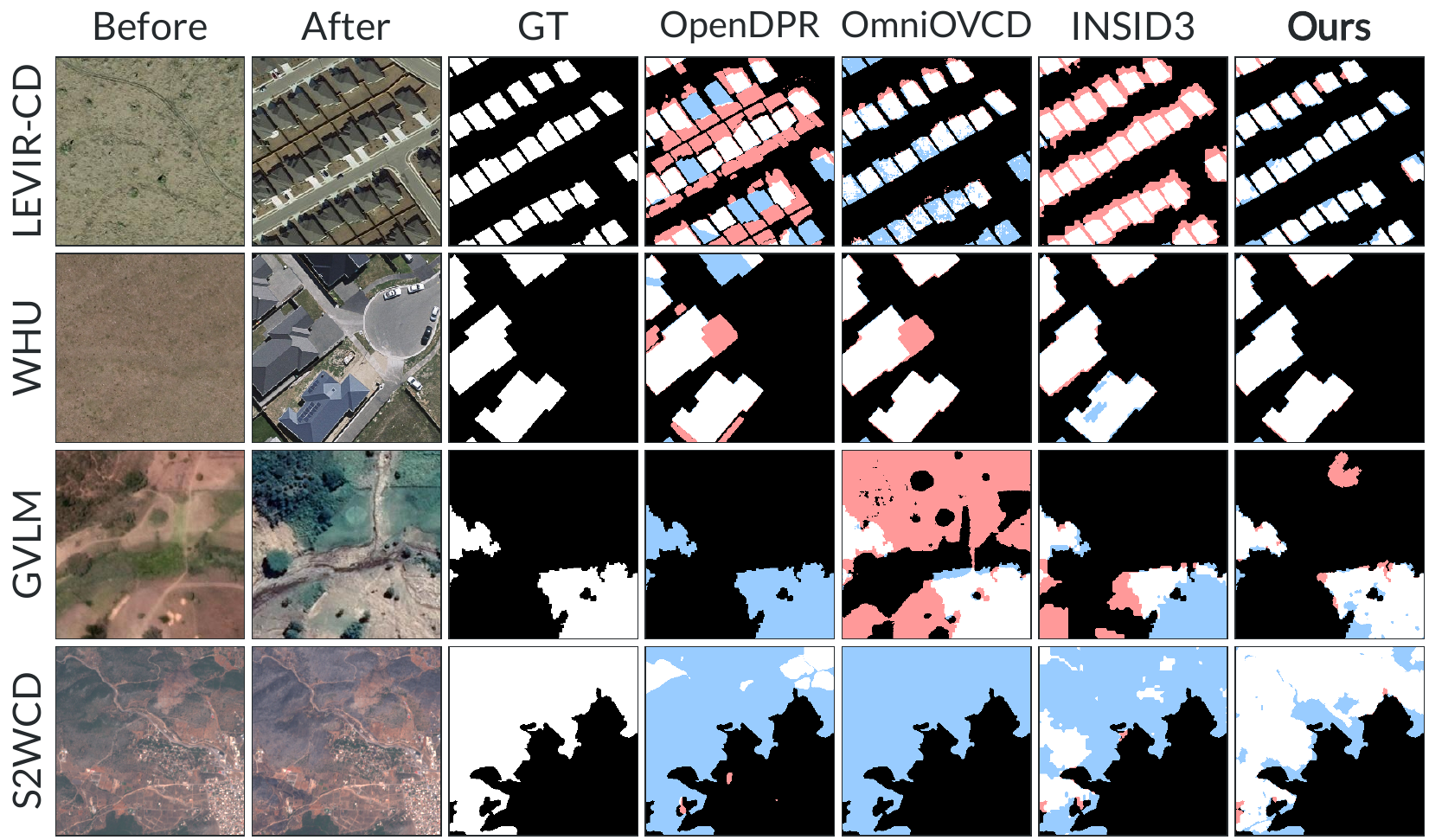}
    \caption{Comparison of building changes (LEVIR-CD, WHU), landslides (GVLM) and wildfire changes (S2WCD). OVTSID3 visualisation was obtained using a text prompt on LEVIR and WHU, and a visual prompt on GVLM and S2WCD (the same example as in INSID3). \textcolor{falsePos}{False positives} are red and \textcolor{falseNeg}{false negatives} blue.}
    \label{fig:qual}
\end{figure}

The qualitative examples presented in \Cref{fig:qual} illustrate that our method yields fewer spurious detections and more complete regions of change, particularly for the displayed natural-disaster scenes. 

\begin{table}[t]
    \centering
    \caption{Ablation of the shared components (F1). Text prompts on LEVIR and WHU, visual prompts on GVLM and S2WCD. Numbers are relative to Ours. (full combinations in Supplementary).}
    \label{tab:abl_shared}
    \setlength{\tabcolsep}{3pt}
    \resizebox{\linewidth}{!}{
    \begin{tabular}{lccccc}
    \toprule
        & LEVIR & WHU & GVLM & S2WCD & \textit{Avg} \\
    \midrule
    \rowcolor{oursblue}
        \textbf{Ours} (clusters \& change map) & \textbf{83.0} & 80.8 & \textbf{55.3} & 73.4 & \textbf{73.1} \\
   w/o positional debiasing & \dneg{1.4} & \dneg{0.3} & \dneg{16.5} & \dneg{21.9} & \dneg{10.0} \\
        Concept mask only (no $C$) & \dneg{18.6} & \dneg{50.4} & \dneg{10.3} & \dneg{14.1} & \dneg{23.3} \\
        Change map only (no prompt) & \dneg{57.4} & \dneg{61.8} & \dneg{23.9} & \dpos{6.8} & \dneg{34.0} \\
        No clusters & \dneg{1.1} & 0.0 & \dpos{2.1} & \dneg{4.4} & \dneg{0.8} \\
        + seed clusters~\cite{cuttano2026insid3} & 0.0 & \dpos{0.8} & \dneg{0.8} & \dpos{0.1} & 0.0 \\
    
    \bottomrule
    \end{tabular}}
\end{table}

\begin{figure}[t]
    \centering
    \includegraphics[width=1\linewidth]{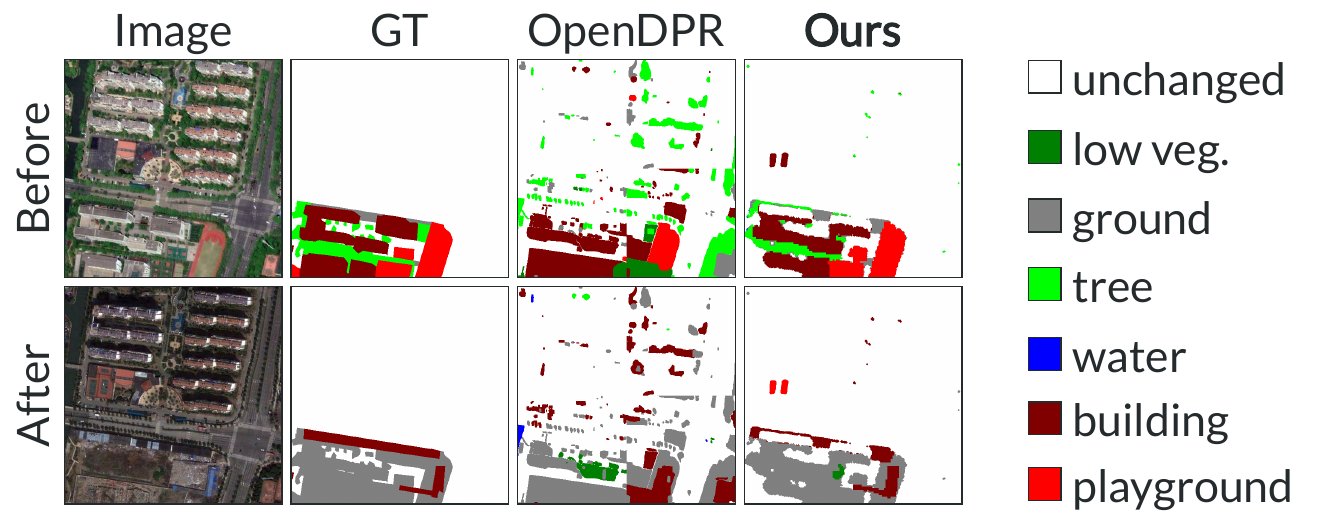}
    \caption{Qualitative comparison of SCD on SECOND. 
    }
    \label{fig:scd_qual}
\end{figure}

\noindent\textbf{Semantic change detection.}
On SECOND, OVTSID3 achieves $30.2\%$ mIoU and $45.1\%$ mean F1 (\Cref{tab:sota_scd}), improving over OpenDPR by $1.6$ and $1.4$ points, respectively. It also exceeds OmniOVCD, which only supports per-class inference, by $3.1$ mIoU and $3.3$ mean F1 points. OVTSID3 obtains the highest scores for classes \textit{water}, \textit{ground}, \textit{building}, and \textit{playground}. \Cref{fig:scd_qual} illustrates the resulting paired semantic maps, showing that our method segments more complete building regions and yields fewer isolated false-positive predictions.

\begin{figure}[t]
    \centering
    \includegraphics[width=1\linewidth]{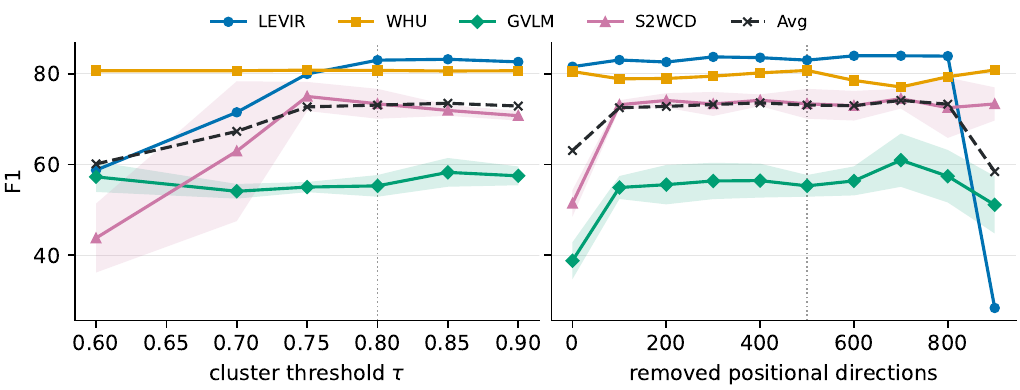}
    \caption{Sensitivity to the cluster threshold $\tau$ (left) and the number of removed positional directions $r$ (right) in average F1. Bands: std over visual references. Dashed vertical: our setting.}
    \label{fig:tau_grid}
\end{figure}

\begin{table*}[t]
    \centering
        \caption{\textbf{Comparison of open-vocabulary methods on the semantic change detection dataset SECOND.} $^\dagger$ denotes per-class inference, \ie, each class is predicted in a separate pass; all other methods predict all classes jointly. M-C-I~\cite{li2025dynearth} is extended to multi-class inference by~\cite{guo2026opendpr} with a single (SP) or multiple (MP) text prompts per class. Results of other methods are taken from~\cite{guo2026opendpr} except for OmniOVCD, which is from their paper. Best and second-best scores are marked.}
    \label{tab:sota_scd}
    \setlength{\tabcolsep}{3pt}
    \begin{tabular}{lcccccccccccccc}
    \toprule
    	\multirow{2}{*}{Method} & \multicolumn{2}{c}{Water} & \multicolumn{2}{c}{Ground} & \multicolumn{2}{c}{Low vegetation} & \multicolumn{2}{c}{Tree} & \multicolumn{2}{c}{Building} & \multicolumn{2}{c}{Playground} & \multicolumn{2}{c}{\textit{Average}}\\
~ & IoU & F1 & IoU & F1 & IoU & F1 & IoU & F1 & IoU & F1 & IoU & F1 & mIoU & mF1\\
 \midrule
I-M-C$^\dagger$~\cite{li2025dynearth}\tiny{AAAI26} & 12.2 & 21.7 & 0.1 & 0.2 & 1.4 & 2.7 & 14.1 & 24.8 & 28.1 & 43.9 & 16.0 & 27.6 & 12.0 & 20.1\\
M-C-I$^\dagger$~\cite{li2025dynearth}\tiny{AAAI26} & 14.3 & 25.1 & 26.2 & 41.6 & \bm2{24.1} & \bm2{38.9} & \bm2{20.3} & \bm2{33.8} & 38.1 & 55.2 & 20.0 & 33.3 & 23.8 & 38.0\\
M-C-I (SP)~\cite{li2025dynearth}\tiny{AAAI26} & 8.4 & 15.5 & 24.3 & 39.1 & 9.4 & 17.2 & 9.0 & 16.5 & 5.1 & 9.8 & 0.2 & 0.4 & 9.4 & 16.4\\
M-C-I (MP)~\cite{li2025dynearth}\tiny{AAAI26} & 10.9 & 19.6 & 25.4 & 40.5 & 22.5 & 36.7 & \bm2{20.3} & \bm2{33.8} & 30.8 & 47.1 & 23.3 & 37.8 & 22.2 & 35.9\\
OpenDPR$^\dagger$~\cite{guo2026opendpr}\tiny{CVPR26} & 12.0 & 21.4 & \bm2{31.2} & \bm2{47.5} & 23.9 & 38.6 & 19.7 & 33.0 & 43.2 & 60.4 & 26.5 & 42.0 & 26.1 & 40.5\\
OpenDPR~\cite{guo2026opendpr}\tiny{CVPR26} & 17.5 & 29.8 & 30.2 & 46.4 & 23.0 & 37.4 & \bm1{20.9} & \bm1{34.5} & 42.4 & 59.5 & \bm2{37.3} & \bm2{54.3} & \bm2{28.6} & \bm2{43.7}\\
OmniOVCD$^\dagger$~\cite{zhang2026omniovcd}\tiny{arXiv26} & \bm2{21.2} & \bm2{35.0} & 27.7 & 43.4 & \bm1{24.5} & \bm1{39.3} & 16.7 & 28.6 & \bm2{45.2} & \bm2{62.3} & 27.0 & 42.4 & 27.1 & 41.8\\
 \midrule
\rowcolor{oursblue}
\textbf{OVTSID3} & \bm1{22.9} & \bm1{37.2} & \bm1{32.8} & \bm1{49.4} & 19.4 & 32.6 & 16.9 & 28.9 & \bm1{49.6} & \bm1{66.3} & \bm1{39.5} & \bm1{56.6} & \bm1{30.2} & \bm1{45.1}\\

    \bottomrule
    \end{tabular}
\end{table*}

\subsection{Ablation Study}
\label{sec:ablations}
We examine the shared pipeline, prompting modalities, and semantic extension. Binary ablations use text on the building datasets (LEVIR \& WHU) and visual references on GVLM and S2WCD unless stated otherwise.

\noindent\textbf{Core ablations.}
\Cref{tab:abl_shared} shows that removing positional debiasing lowers the average by 10 points, with the largest reductions on GVLM (16.5) and S2WCD (21.9). These results support the use of debiased features~\cite{cuttano2026insid3} in other remote sensing tasks, since positional bias is especially detrimental in RS, as there is no canonical orientation~\cite{rolf2024satellite}.

Results show that both masks are needed: the concept mask alone achieves an average F1 score 23.3 points lower, and the change map 34 points lower than their combination. Without a prompt, the change map detects nearly all changes (recall ${>}97\%$ on LEVIR-CD and GVLM) but cannot tell apart the target from other changes (precision ${<}19\%$). The exception is S2WCD, a low-resolution dataset where burn scars are almost the only change: the change map alone achieves very good results, and the prompt raises precision, but sacrifices recall. 

Removing clusters slightly reduces the average performance. We also observe that using the seed cluster, as in INSID3, instead of the cluster score alone (\Cref{sec:concept_mask}) does not improve CD results, making our method simpler.

\noindent\textbf{Sensitivity.}
\Cref{fig:tau_grid} varies the cluster threshold $\tau$ and the number of removed positional directions $r$. Results are stable for $\tau\in[0.75,0.9]$, which demonstrates the need for more granular clustering in the RS domain, where INSID3's default of $0.6$ fails. The exact degree of debiasing $r$ is not critical except at the extremes; however, results are substantially worse without it.

\noindent\textbf{Visual scoring and prototype construction.}
Reference-based thresholding and paired-score centring improve average F1 across the disaster datasets (\Cref{tab:abl_vis}). The full configuration achieves $64.4\%$ F1, outperforming fixed thresholding by $7.2$ points and the variant without centring by $2.3$ points. On S2WCD, however, both alternatives achieve higher F1, primarily through increased recall. This is particularly beneficial given the dataset's large proportion of changed pixels. Nevertheless, these gains come at the expense of substantially worse performance on GVLM, highlighting the limited generalizability of these settings and the advantage of our adaptive formulation. Finally, constructing prototypes from the most centroid-similar reference locations yields a modest $0.4$-point improvement over averaging the entire annotated region like INSID3.

\begin{table}[t]
    \centering
    \caption{Visual prompt components (F1). Numbers are relative to Ours.}
    \label{tab:abl_vis}
    \begin{tabular}{lccc}
    \toprule
        & GVLM & S2WCD & \textit{Avg} \\
    \midrule
    \rowcolor{oursblue}
        \textbf{Ours} & \textbf{55.3} & 73.4 & \textbf{64.4} \\
        Fixed $\delta=0.01$ & \dneg{19.3} & \dpos{5.1} & \dneg{7.2} \\
        w/o pair centring & \dneg{11.1} & \dpos{6.5} & \dneg{2.3} \\
        Mask mean & \dneg{2.3} & \dpos{1.5} & \dneg{0.4} \\
    \bottomrule
    \end{tabular}
\end{table}

\noindent\textbf{Semantic components.}
All tested components improve SECOND results (\Cref{tab:abl_scd}). Omitting the class-change condition lowers the mean F1 score by 1.8 points, indicating benefit from an explicit requirement that the predicted classes change. However, removing the class-agnostic change map reduces the results much more, by 6.9 mF1. Removing positional debiasing reduces scores by 6.3. Direct classification, instead of cluster-level assignment, and replacing top-$k$ description averaging with simple argmax both reduce results, though less than other elements.

\begin{table}[t]
    \centering
    \caption{Ablation on semantic change detection (SECOND). Numbers are relative to Ours.}
    \label{tab:abl_scd}
    \begin{tabular}{lcc}
    \toprule
        & mIoU & mF1 \\
    \midrule
    \rowcolor{oursblue}
        \textbf{Ours} & \textbf{30.2} & \textbf{45.1} \\
        w/o class gate ($y_1\neq y_2$) & \dneg{1.8} & \dneg{1.8} \\
        w/o change map & \dneg{6.0} & \dneg{6.9} \\
        w/o positional debiasing & \dneg{5.0} & \dneg{6.3} \\
        w/o clusters & \dneg{0.5} & \dneg{0.5} \\
        Max instead of top-$k$ descriptions & \dneg{0.7} & \dneg{0.7} \\
    \bottomrule
    \end{tabular}
\end{table}

\noindent\textbf{Efficiency.}
On an A100 GPU (protocol in Supplementary), OVTSID3 processes a $1024\times1024$ image pair in $779$\,ms with a text prompt and $1374$\,ms with a visual prompt, using $2.2$\,GB of peak memory. INSID3 needs $1191$\,ms and $2.3$\,GB, and OmniOVCD $3359$\,ms and $5.3$\,GB with the prompts used in \Cref{tab:sota_ovcd}.

\noindent\textbf{Limitations and future work.}
Prompt choice remains important: a single non-representative visual reference can transfer poorly across building appearances, while text is less effective for the evaluated landslides. The lower SECOND scores for low vegetation and trees also indicate remaining semantic ambiguity in RGB-only VLMs.

\section{Conclusion}

We presented OVTSID3, a training-free framework that unifies spatial grouping, concept identification, and temporal comparison within a single frozen DINOv3 representation. It supports both zero-shot textual and one-shot visual prompting through a shared inference pipeline, enabling flexible concept specification without task-specific training. OVTSID3 achieves the highest average F1 among the evaluated binary change detection methods and extends to semantic change detection on SECOND. By allowing concepts to be specified at inference time in either textual or visual formats, our approach represents a step towards reusable perception tools for agentic Earth observation systems, where detection targets, as well as their modality, may change dynamically in response to user requests.

\section*{LLM Usage Statement}
We used ChatGPT (OpenAI) and Claude (Anthropic) as author-directed assistive tools for language refinement, code refinement, help with experimental planning and analysis, figure and diagram refinement,
and refinement of research ideas. We also provided the manuscript to these tools for reviewer-style feedback. No LLMs or other generative AI tools were used for reference generation, synthetic data generation, dataset creation, or annotation. All tool-assisted content was reviewed, substantially revised, and verified by the authors, who take full responsibility
for the final manuscript and all scientific claims.

{
    \small
    \bibliographystyle{ieeenat_fullname}
    \bibliography{main}
}


\renewcommand{\thesection}{\Alph{section}}

\renewcommand{\thetable}{S\arabic{table}}

\renewcommand{\thefigure}{S\arabic{figure}}
\setcounter{table}{0} 
\setcounter{figure}{0} 
\appendix
\maketitlesupplementary

In these supplementary materials, we provide additional details that extend beyond the scope of the main manuscript. It is organised as follows: 
\begin{itemize}
    \item \textbf{Societal impact} in \Cref{sec:societal}
    \item \textbf{Extended dataset details} in Section~\ref{a:data}.
    \item \textbf{Extended results} with standard deviation (main and ablation results) and hyperparameter validation in Section~\ref{a:ext_res}.
    \item \textbf{Additional qualitative results}, including failure cases and more SCD results in Section~\ref{a:qual}.
    \item \textbf{Extended implementation details} for the model, parameters, prompts, and related methods in Section~\ref{a:impl}.
    \item \textbf{Computational efficiency} protocol and extended results in Section~\ref{a:eff}.

\end{itemize}

\section{Potential negative societal impact}
\label{sec:societal}

We strongly believe that our method has significant potential for socially beneficial applications, particularly in the context of natural disasters. However, its general-purpose nature and planned public release of code also introduce risks of misuse, including for surveillance and military applications. While we do not endorse such uses, we acknowledge that they cannot be fully prevented.

Another potentially negative impact is the compute required. While OVTSID3 operates in a training-free manner, it still requires a decent amount of compute. This essentially means that for large-scale operation, environmental factors should be considered.

\subsection{Used compute}

All experiments for this paper, including development, ablations, and multiple reference seeds, used about 1,000 GPU-hours for our method and about 150 GPU-hours for the baselines, across a mix of NVIDIA L4, A100, H100, B200, and B300 GPUs.

\section{Extended dataset details}
\label{a:data}

\begin{table*}[]
\centering
    \caption{Additional details for the datasets used in the paper.}

\resizebox{\linewidth}{!}{
\setlength{\tabcolsep}{3pt}

    \begin{tabular}{lccccccccc}
        \toprule
 & Acquisition & Resolution & Change Type & Interval & Region & \makecell{Image count \\ train\textbackslash val\textbackslash test} & Patch & \makecell{Changed \\ Pixels} & \makecell{Unchanged \\ Pixels} \\
\midrule
LEVIR~\cite{chen2020levirStanet} & \makecell{Google Earth \\ satellite} & 0.5m & Building & 2002-2018 & \makecell{20 regions \\ in US} & \makecell{7120\\1024\\2048} & $256\times256$ & $4.7~\%$ & $95.3~\%$ \\\midrule
WHU~\cite{ji2018whu} & Aerial & 0.3m & Building & 2012-2016 & \makecell{Christchurch,\\ New Zealand} & \makecell{4980\\-\\2640} & $256\times256$ & $4.3~\%$ & $95.7~\%$ \\\midrule
GVLM~\cite{zhang2023gvlm} & \makecell{Google Earth \\ satellite \\ (SPOT-6)} & 0.59m & Landslide & 2010-2021 & \makecell{17 regions\\ worldwide} & \makecell{4558\\1519\\1519} & $256\times256$ & $6.6~\%$ & $93.4~\%$ \\\midrule
S2WCD~\cite{valsamis2024s2wcd} & \makecell{Sentinel-2 \\ satellite (RGB)} & 10m & Wildfire & \makecell{2021-2023} & \makecell{25 fire event \\ S.\ Europe, Australia} & \makecell{336\\-\\320} & $256\times256$ & $35.6~\%$ & $64.4~\%$ \\\midrule
SECOND~\cite{yang2020second} & \makecell{Aerial \\ (multiple sensors)} & undisclosed & \makecell{Semantic dataset:\\ground, tree, low vegetation\\ water, buildings\\ playgrounds} & undisclosed & \makecell{Hangzhou, Chengdu,\\Shanghai, China} & \makecell{2968\\-\\1694} & $512\times512$ & 19.1 \% & 80.9 \% \\
\bottomrule
    \end{tabular}
    }
    \label{atab:data_extend}
\end{table*}

More dataset details are provided in \Cref{atab:data_extend}. Our benchmarks span two building-change datasets and two natural-disaster datasets for landslides and wildfires. We also include the SECOND semantic change detection dataset, which has 6 land cover classes. The datasets vary in ground sampling distance (GSD), acquisition sensor, and scale, ranging from a few hundred to several thousand image pairs. Many previous OVCD methods were evaluated only on buildings and SECOND, so this diversity strengthens the robustness and generality of our conclusions. 

\subsection{Data implementation details}
\label{asub:data_impl}

\textbf{Dataset splits.} Official train and test splits are used for LEVIR and SECOND. For WHU, we process the data ourselves but adhere to the author-provided train and test regions to ensure compatibility. GVLM has no official split, but we use the publicly available splits from HuggingFace. Since S2WCD does not have public splits, we split it approximately 50/50 ourselves to ensure no event overlap; the split events are below. We also use a validation set (for optimal threshold and visual prototype $q$ selection) from LEVIR and GVLM.

\begin{itemize}
    \item LEVIR: \textcolor{weights}{HF ericyu: LEVIRCD\_Cropped256}
    \item WHU: \textcolor{weights}{official web page}
    \item GVLM: \textcolor{weights}{HF ericyu: GVLM\_Cropped\_256}
    \item S2WCD: \textcolor{weights}{official IEEE dataport page.}
\end{itemize}
SECOND was obtained via GitHub \textcolor{weights}{captain-whu: SCD}.

For S2WCD, we split the 41 tiles into training and test sets by fire event, so that no event appears in both splits. Tiles of the same event (e.g., Evros~1--4) share the fire and the acquisition dates and are kept together.
The training set contains 21 tiles from 11 events: Attica, Carrascal, Ebo, Evros (4), Folgoso (4), Manh, Monaco,
Murca, North Euvoia (2), Obarco, and Rhodes (4).
The test set contains 20 tiles from 14 events: Almonaster, Australia (2), Bejis, Caceres, Cohilva (2), Jubrique,
Landes, Los Guajares, Navvara, Olympia (4), Sardinia, Tenerife (2), W.~Attica, and Zaragosa.
Numbers in parentheses denote the number of tiles per event. The number in \Cref{atab:data_extend} is the number of cropped $256\times256$ patches.

\section{Extended and additional results}
\label{a:ext_res}

This section contains results with standard deviation for main results and ablations, as well as hyperparameter selection results.

\subsection{Full results with std}

Full results with standard deviation are included in \Cref{tab:supp_sota_ovcd}. Here, we also include I3PE~\cite{chen2023_i3pe} and AdaptOVCD~\cite{dou2026adaptovcd}, which were omitted from the main paper due to space constraints and because their results do not alter the ranking. With its default settings, AdaptOVCD predicts almost no changes on S2WCD (0.0 F1), indicating the limits of SAM-based methods on low-resolution imagery.

\begin{table*}[!ht]
    \centering
    \caption{Change detection results on building changes (LEVIR-CD and WHU), landslides (GVLM), and wildfire changes (S2WCD). Prompt: a text prompt or a single masked visual reference (same references used for all visual-prompt methods). \textit{Avg} denotes the mean over all four datasets. The \colorbox{gold}{best} and \colorbox{silver}{second-best} results are highlighted. Standard deviation is reported over 3 random seeds for non-deterministic runs (visual or training-based UCD). $*$ indicates the method was trained on in-domain data.}
    \resizebox{\linewidth}{!}{
\setlength{\tabcolsep}{3pt}
    \begin{tabular}{lccccccccccc}
    \toprule
    	\multirow{2}{*}{~} & \multirow{2}{*}{Prompt} & \multicolumn{2}{c}{LEVIR} & \multicolumn{2}{c}{WHU} & \multicolumn{2}{c}{GVLM} & \multicolumn{2}{c}{S2WCD} & \multicolumn{2}{c}{\textit{Avg}}\\
~ & ~ & IoU & F1 & IoU & F1 & IoU & F1 & IoU & F1 & IoU & F1\\
 \midrule
\multicolumn{12}{l}{\textit{Unsupervised}}\\
Pixel Difference & -- & 4.7 & 8.9 & 7.9 & 14.7 & 13.3 & 23.4 & 24.1 & 38.8 & 12.5 & 21.4\\
I3PE~\cite{chen2023_i3pe}\tiny{ISPRS23} & -- & 9.4{\scriptsize$\pm$0.2} & 17.1{\scriptsize$\pm$0.3} & 2.1{\scriptsize$\pm$0.2} & 4.1{\scriptsize$\pm$0.3} & 12.7{\scriptsize$\pm$0.2} & 22.5{\scriptsize$\pm$0.3} & 1.3{\scriptsize$\pm$0.4} & 2.6{\scriptsize$\pm$0.7} & 6.4 & 11.6\\
S2C (DINOv3)$^*$~\cite{ding2025s2c}\tiny{AAAI26} & -- & 20.2{\scriptsize$\pm$1.0} & 33.7{\scriptsize$\pm$1.5} & 17.3{\scriptsize$\pm$3.4} & 29.4{\scriptsize$\pm$5.0} & 27.1{\scriptsize$\pm$0.6} & 42.7{\scriptsize$\pm$0.7} & 0.0 & 0.0 & 16.2 & 26.4\\
AnyChange~\cite{zheng2024anychange}\tiny{NeurIPS24} & -- & 14.6 & 25.4 & 12.4 & 22.0 & 11.6 & 20.8 & 4.2 & 8.0 & 10.7 & 19.1\\
\quad$\hookrightarrow$ point query & Point & 15.6{\scriptsize$\pm$1.7} & 27.0{\scriptsize$\pm$2.5} & 23.2{\scriptsize$\pm$2.2} & 37.6{\scriptsize$\pm$2.9} & 6.7{\scriptsize$\pm$2.6} & 12.5{\scriptsize$\pm$4.6} & 0.7{\scriptsize$\pm$0.8} & 1.4{\scriptsize$\pm$1.5} & 11.6 & 19.6\\
Mason$^*$~\cite{rolih2026mason}\tiny{TGRS26} & -- & 24.8{\scriptsize$\pm$0.3} & 39.7{\scriptsize$\pm$0.4} & 20.3{\scriptsize$\pm$2.6} & 33.8{\scriptsize$\pm$3.6} & \bm1{38.3}{\scriptsize$\pm$2.7} & \bm1{55.4}{\scriptsize$\pm$2.9} & 33.8{\scriptsize$\pm$6.5} & 50.3{\scriptsize$\pm$7.1} & 29.3 & 44.8\\
 \midrule
\multicolumn{12}{l}{\textit{Supervised open-vocabulary}}\\
\color{gray}Seg2Change~\cite{su2026seg2change}\tiny{ACMM26} & \color{gray}Text & \color{gray}64.9 & \color{gray}78.7 & \color{gray}75.7 & \color{gray}86.2 & \color{gray}44.1 & \color{gray}61.2 & \color{gray}2.4 & \color{gray}4.7 & \color{gray}46.8 & \color{gray}57.7\\
 \midrule
\multicolumn{12}{l}{\textit{In-context segmentation and mask XOR}}\\
INSID3~\cite{cuttano2026insid3}\tiny{CVPR26} & Visual & 27.5{\scriptsize$\pm$1.4} & 43.2{\scriptsize$\pm$1.7} & 36.1{\scriptsize$\pm$3.1} & 53.0{\scriptsize$\pm$3.3} & 22.6{\scriptsize$\pm$3.3} & 36.7{\scriptsize$\pm$4.3} & 25.1{\scriptsize$\pm$3.3} & 40.0{\scriptsize$\pm$4.3} & 27.8 & 43.2\\
\quad$\hookrightarrow$ + our change map & Visual & 45.3{\scriptsize$\pm$4.7} & 62.2{\scriptsize$\pm$4.5} & 52.0{\scriptsize$\pm$4.9} & 68.3{\scriptsize$\pm$4.3} & 38.0{\scriptsize$\pm$2.1} & 55.1{\scriptsize$\pm$2.2} & 55.0{\scriptsize$\pm$13.8} & 70.2{\scriptsize$\pm$12.2} & 47.6 & 64.0\\
 \midrule
\multicolumn{12}{l}{\textit{Training-free open-vocabulary}}\\
SCM~\cite{tan2024scm}\tiny{IGARSS24} & Text & 15.8 & 27.3 & 20.0 & 33.3 & 16.3 & 28.0 & 5.4 & 10.2 & 14.4 & 24.7\\
DynEarth M-C-I~\cite{li2025dynearth}\tiny{AAAI26} & Text & 32.9 & 49.5 & 57.3 & 72.9 & 16.4 & 28.2 & 6.0 & 11.3 & 28.1 & 40.4\\
DynEarth I-M-C~\cite{li2025dynearth}\tiny{AAAI26} & Text & 20.0 & 33.3 & 8.7 & 16.0 & 0.7 & 1.3 & 1.0 & 2.0 & 7.6 & 13.2\\
OpenDPR~\cite{guo2026opendpr}\tiny{CVPR26} & Text & 44.8 & 61.9 & 55.6 & 71.4 & 21.6 & 35.5 & 10.2 & 18.5 & 33.0 & 46.9\\
AdaptOVCD~\cite{dou2026adaptovcd}\tiny{arXiv26} & Text & 51.5 & 68.0 & 62.0 & 76.5 & 6.1 & 11.6 & 0.0 & 0.0 & 29.9 & 39.0\\
OmniOVCD~\cite{zhang2026omniovcd}\tiny{arXiv26} & Text & \bm2{67.8} & \bm2{80.8} & \bm2{65.7} & \bm2{79.3} & 21.9 & 35.9 & 32.0 & 48.5 & 46.9 & 61.1\\
\rowcolor{oursblue}
\textbf{OVTSID3} (visual) & Visual & 41.9{\scriptsize$\pm$12.1} & 58.4{\scriptsize$\pm$12.7} & 61.0{\scriptsize$\pm$9.4} & 75.5{\scriptsize$\pm$7.4} & \bm2{38.2}{\scriptsize$\pm$2.3} & \bm2{55.3}{\scriptsize$\pm$2.4} & \bm1{58.0}{\scriptsize$\pm$4.1} & \bm1{73.4}{\scriptsize$\pm$3.3} & \bm2{49.8} & \bm2{65.6}\\
\rowcolor{oursblue}
\textbf{OVTSID3} (text) & Text & \bm1{71.0} & \bm1{83.0} & \bm1{67.7} & \bm1{80.8} & 28.3 & 44.1 & \bm2{56.2} & \bm2{71.9} & \bm1{55.8} & \bm1{70.0}\\

    \bottomrule
    \end{tabular}
    }
    \label{tab:supp_sota_ovcd}
\end{table*}

\subsection{Ablation results with std}

Full ablation with text prompts is shown in \Cref{tab:supp_abl_shared_txt} and for visual prompts in \Cref{tab:supp_abl_shared_vis}. 

\begin{table}[h]
    \centering
    \caption{Core ablations for text-prompted version.}
    \setlength{\tabcolsep}{3pt}
    \label{tab:supp_abl_shared_txt}
    \resizebox{\linewidth}{!}{
    \
    \begin{tabular}{lccccc}
    \toprule
        & LEVIR & WHU & GVLM & S2WCD & \textit{Avg} \\
    \midrule
        \textbf{Ours} & 83.0 & 80.8 & 44.1 & 71.9 & 70.0 \\
        Concept mask only (no $C$) & 64.4 & 30.4 & 28.9 & 59.2 & 45.7 \\
        w/o positional debiasing & 81.6 & 80.5 & 31.9 & 6.9 & 50.2 \\
        No clusters & 81.9 & 80.8 & 43.2 & 68.7 & 68.6 \\
        + seed clusters~\cite{cuttano2026insid3} & 83.0 & 81.6 & 40.0 & 74.6 & 69.8 \\
        XOR instead of union & 83.0 & 86.0 & 43.0 & 47.7 & 64.9 \\
        Change map $C$ only (no prompt) & 25.6 & 19.0 & 31.4 & 80.2 & 39.1 \\
        Target similarity only & 72.1 & 70.3 & 30.7 & 62.9 & 59.0 \\
    \bottomrule
    \end{tabular}}
\end{table}

\begin{table}[h]
    \centering
    \caption{Core ablation for the visually prompted version.}
    \setlength{\tabcolsep}{3pt}
    \label{tab:supp_abl_shared_vis}
    \resizebox{\linewidth}{!}{
    \begin{tabular}{lccccc}
    \toprule
        & LEVIR & WHU & GVLM & S2WCD & \textit{Avg} \\
    \midrule
        \textbf{Ours} & 58.4{\scriptsize$\pm$12.7} & 75.5{\scriptsize$\pm$7.4} & 55.3{\scriptsize$\pm$2.4} & 73.4{\scriptsize$\pm$3.3} & 65.6 \\
        Concept mask only (no $C$) & 41.2{\scriptsize$\pm$20.1} & 33.3{\scriptsize$\pm$5.1} & 45.0{\scriptsize$\pm$6.9} & 59.3{\scriptsize$\pm$0.5} & 44.7 \\
        w/o positional debiasing & 53.3{\scriptsize$\pm$20.9} & 60.6{\scriptsize$\pm$4.2} & 38.8{\scriptsize$\pm$4.1} & 51.5{\scriptsize$\pm$2.9} & 51.0 \\
        No clusters & 45.4{\scriptsize$\pm$19.7} & 70.8{\scriptsize$\pm$15.3} & 57.4{\scriptsize$\pm$1.7} & 69.0{\scriptsize$\pm$0.9} & 60.7 \\
        + seed clusters~\cite{cuttano2026insid3} & 51.0{\scriptsize$\pm$23.5} & 67.5{\scriptsize$\pm$9.8} & 54.5{\scriptsize$\pm$2.8} & 73.5{\scriptsize$\pm$2.9} & 61.6 \\
        XOR instead of union & 43.8{\scriptsize$\pm$20.5} & 71.2{\scriptsize$\pm$15.6} & 54.7{\scriptsize$\pm$2.9} & 46.1{\scriptsize$\pm$6.2} & 53.9 \\
        Change map $C$ only (no prompt) & 25.6 & 19.0 & 31.4 & 80.2 & 39.1 \\
    \bottomrule
    \end{tabular}}
\end{table}

\Cref{tab:supp_abl_vis} contains the full ablation of options for visual prompting.

\begin{table}[h]
    \centering
    \caption{Ablation of visual prompting building blocks.}
    \setlength{\tabcolsep}{3pt}
    \label{tab:supp_abl_vis}
    \resizebox{\linewidth}{!}{
    \begin{tabular}{lccccc}
    \toprule
        & LEVIR & WHU & GVLM & S2WCD & \textit{Avg} \\
    \midrule
        \textbf{Ours} & 58.4{\scriptsize$\pm$12.7} & 75.5{\scriptsize$\pm$7.4} & 55.3{\scriptsize$\pm$2.4} & 73.4{\scriptsize$\pm$3.3} & 65.6 \\
        Fixed $\delta=0.01$ & 30.7{\scriptsize$\pm$0.5} & 29.2{\scriptsize$\pm$0.5} & 36.0{\scriptsize$\pm$0.5} & 78.5{\scriptsize$\pm$0.2} & 43.6 \\
        w/o pair centring & 51.8{\scriptsize$\pm$21.0} & 52.5{\scriptsize$\pm$17.7} & 44.2{\scriptsize$\pm$5.9} & 79.9{\scriptsize$\pm$1.9} & 57.1 \\
        Mask mean & 57.7{\scriptsize$\pm$13.9} & 70.0{\scriptsize$\pm$9.9} & 53.0{\scriptsize$\pm$1.5} & 74.9{\scriptsize$\pm$2.5} & 63.9 \\
    \bottomrule
    \end{tabular}}
\end{table}

\Cref{tab:supp_abl_scd} contains results for all SCD ablations.

\begin{table*}[t]
    \centering
    \caption{SCD ablation; the best result is bolded for easier viewing.}
    \setlength{\tabcolsep}{3pt}
    \label{tab:supp_abl_scd}
    \resizebox{\linewidth}{!}{
    \begin{tabular}{lcccccccccccccc}
    \toprule
        & \multicolumn{2}{c}{Low veg.} & \multicolumn{2}{c}{Ground} & \multicolumn{2}{c}{Tree} & \multicolumn{2}{c}{Water} & \multicolumn{2}{c}{Building} & \multicolumn{2}{c}{Playground} & \multicolumn{2}{c}{Mean} \\
        \cmidrule(lr){2-3} \cmidrule(lr){4-5} \cmidrule(lr){6-7} \cmidrule(lr){8-9} \cmidrule(lr){10-11} \cmidrule(lr){12-13} \cmidrule(lr){14-15}
        & IoU & F1 & IoU & F1 & IoU & F1 & IoU & F1 & IoU & F1 & IoU & F1 & IoU & F1 \\
    \midrule
        \textbf{Ours} & 19.4 & 32.6 & \textbf{32.8} & \textbf{49.4} & 16.9 & 28.9 & 22.9 & 37.2 & \textbf{49.6} & \textbf{66.3} & \textbf{39.5} & \textbf{56.6} & \textbf{30.2} & \textbf{45.1} \\
        w/o change map & \textbf{20.3} & \textbf{33.7} & 26.6 & 42.0 & 15.2 & 26.3 & 14.5 & 25.3 & 37.9 & 54.9 & 30.7 & 47.0 & 24.2 & 38.2 \\
        w/o class gate ($y_1\neq y_2$) & 19.5 & 32.6 & 27.4 & 43.0 & \textbf{17.1} & \textbf{29.2} & \textbf{23.1} & \textbf{37.6} & 45.0 & 62.0 & 38.5 & 55.6 & 28.4 & 43.3 \\
        w/o positional debiasing & 12.6 & 22.4 & 29.8 & 45.9 & 11.3 & 20.3 & 17.4 & 29.7 & 44.7 & 61.8 & 35.5 & 52.4 & 25.2 & 38.8 \\
        w/o clusters & 19.4 & 32.5 & 32.4 & 49.0 & 16.9 & 29.0 & 22.7 & 37.0 & 49.5 & 66.2 & 37.2 & 54.2 & 29.7 & 44.6 \\
        Max instead of top-$k$ descriptions & 19.9 & 33.2 & 32.5 & 49.1 & 16.0 & 27.6 & 23.0 & 37.4 & 48.7 & 65.5 & 36.8 & 53.8 & 29.5 & 44.4 \\
    \bottomrule
    \end{tabular}}
\end{table*}

\subsection{Hyperparameter Validation}
\label{supp:val}

We check the text threshold $\delta$ and the cluster threshold $\tau$ on the LEVIR-CD validation set and the
prototype fraction $q$ on the GVLM validation set
(\Cref{tab:supp_val}). Visual results are from three seed repeats.

\paragraph{Text threshold $\delta$.}
Since the text score is a foreground--background margin, its natural decision point is zero. Validation F1 is
flat for $\delta\in[0,0.01]$ ($81.5$--$82.1\%$) and peaks at $\delta=0.01$, which we use for all datasets. Larger
values quickly discard true target clusters: F1 drops to $75.2\%$ at $\delta=0.02$ and collapses at $\delta=0.05$
($6.6\%$).

\paragraph{Cluster threshold $\tau$.}
Validation F1 is flat for $\tau\in[0.75,0.9]$ ($81.2$--$82.1\%$) and peaks at $\tau=0.8$, which we use throughout.
Lower thresholds merge clusters across object boundaries: F1 drops to $74.8\%$ at $\tau=0.7$ and $64.8\%$ at
$\tau=0.6$.

\paragraph{Prototype fraction $q$.}
Any fraction $q\in[0.05,0.5]$ gives $53.5$--$53.9\%$ F1, so the exact value is not critical; we use $q=0.25$
from this plateau. Pooling the whole reference mask ($q=1$) lowers F1 to $51.7\%$ and increases the variance across
references, as the mask also covers locations that do not belong to the concept (e.g., unchanged roads inside a
burnt area).

\begin{table*}[h]
    \centering
    \caption{Hyperparameter validation (F1). Bold: our setting.}
    \label{tab:supp_val}
    \begin{tabular}{lcccccc}
    \toprule
        $\delta$ (text) & 0 & 0.005 & \textbf{0.01} & 0.02 & 0.05 \\
        LEVIR-CD val & 81.5 & 81.7 & \textbf{82.1} & 75.2 & 6.6 \\
    \midrule
        $\tau$ (text) & 0.6 & 0.7 & 0.75 & \textbf{0.8} & 0.85 & 0.9 \\
        LEVIR-CD val & 64.8 & 74.8 & 81.2 & \textbf{82.1} & 81.8 & 81.3 \\
    \midrule
        $q$ (visual) & 0.05 & 0.1 & \textbf{0.25} & 0.5 & 1.0 \\
        GVLM val & 53.5{\scriptsize$\pm$1.0} & 53.7{\scriptsize$\pm$1.3} & \textbf{53.8{\scriptsize$\pm$1.3}} & 53.9{\scriptsize$\pm$1.3} & 51.7{\scriptsize$\pm$2.7} \\
    \bottomrule
    \end{tabular}
\end{table*}

\section{Additional qualitative results}
\label{a:qual}

\begin{figure}
    \centering
    \includegraphics[width=1\linewidth]{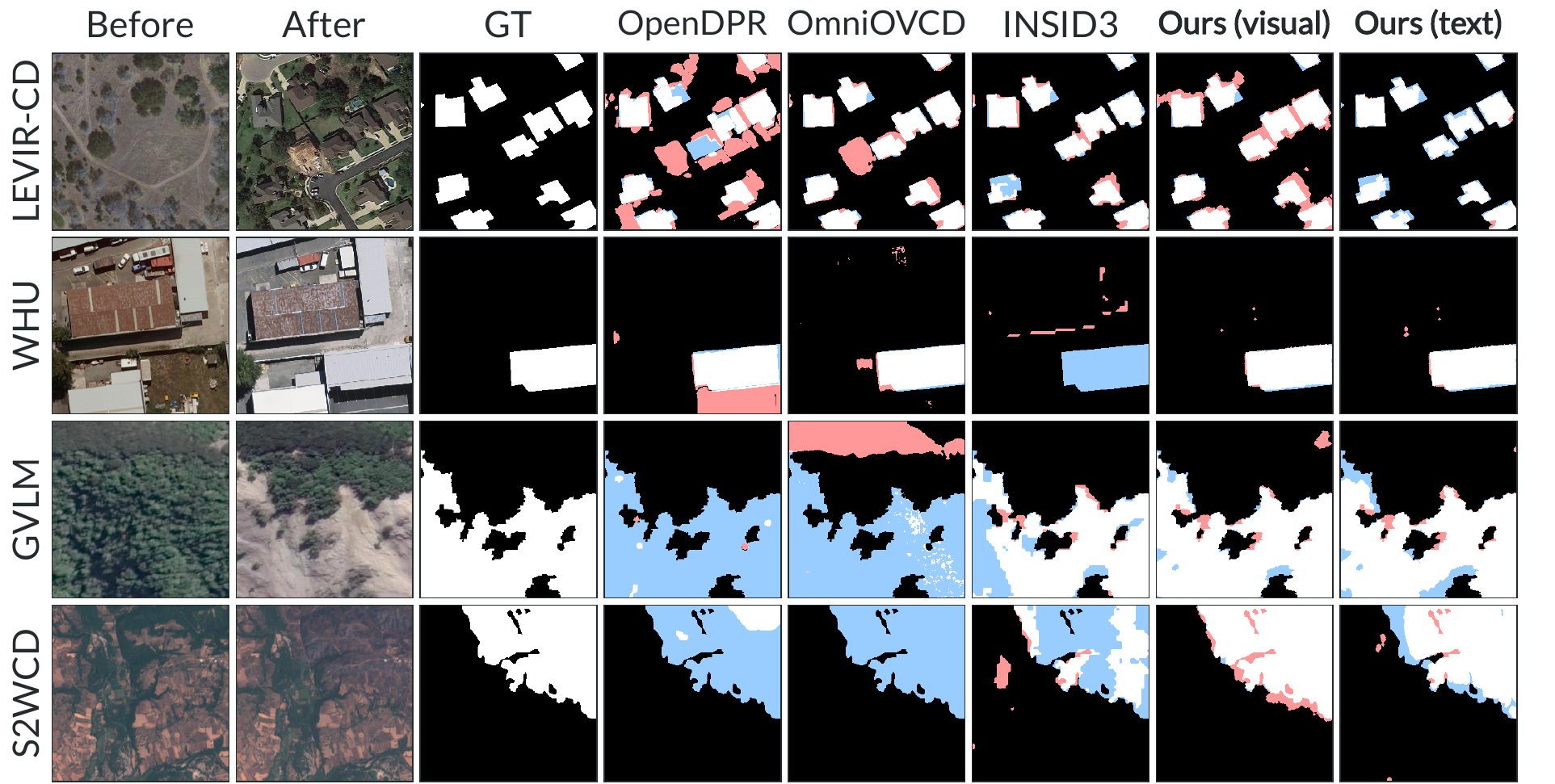}
    \caption{Comparison of building changes (LEVIR-CD, WHU), landslides (GVLM) and wildfire changes (S2WCD). OVTSID3 visualisation was obtained using a text prompt and a visual prompt (the same example as in INSID3). \textcolor{falsePos}{False positives} are red and \textcolor{falseNeg}{false negatives} blue.}
    \label{fig:a_qual}
\end{figure}

Additional BCD qualitative results are presented in \Cref{fig:a_qual}.

\begin{figure}
    \centering
    \includegraphics[width=1\linewidth]{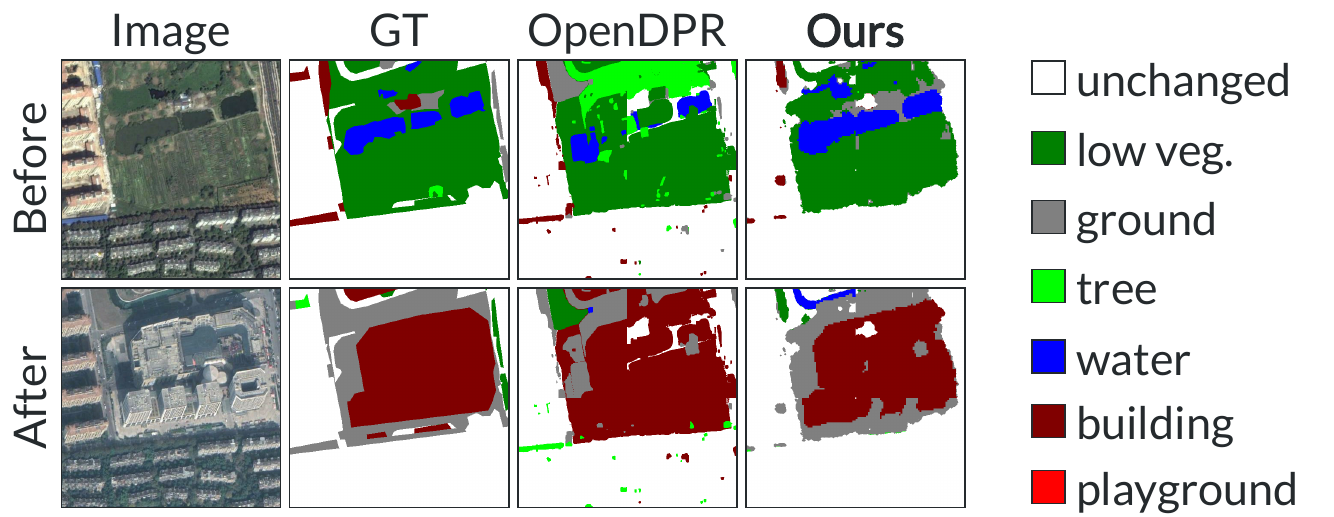}
    \caption{Qualitative comparison of SCD on SECOND.}
    \label{fig:a_scd}
\end{figure}

Additional SCD qualitative results are presented in \Cref{fig:a_scd}.

\subsection{Failure case analysis}

\begin{figure}
    \centering
    \includegraphics[width=1\linewidth]{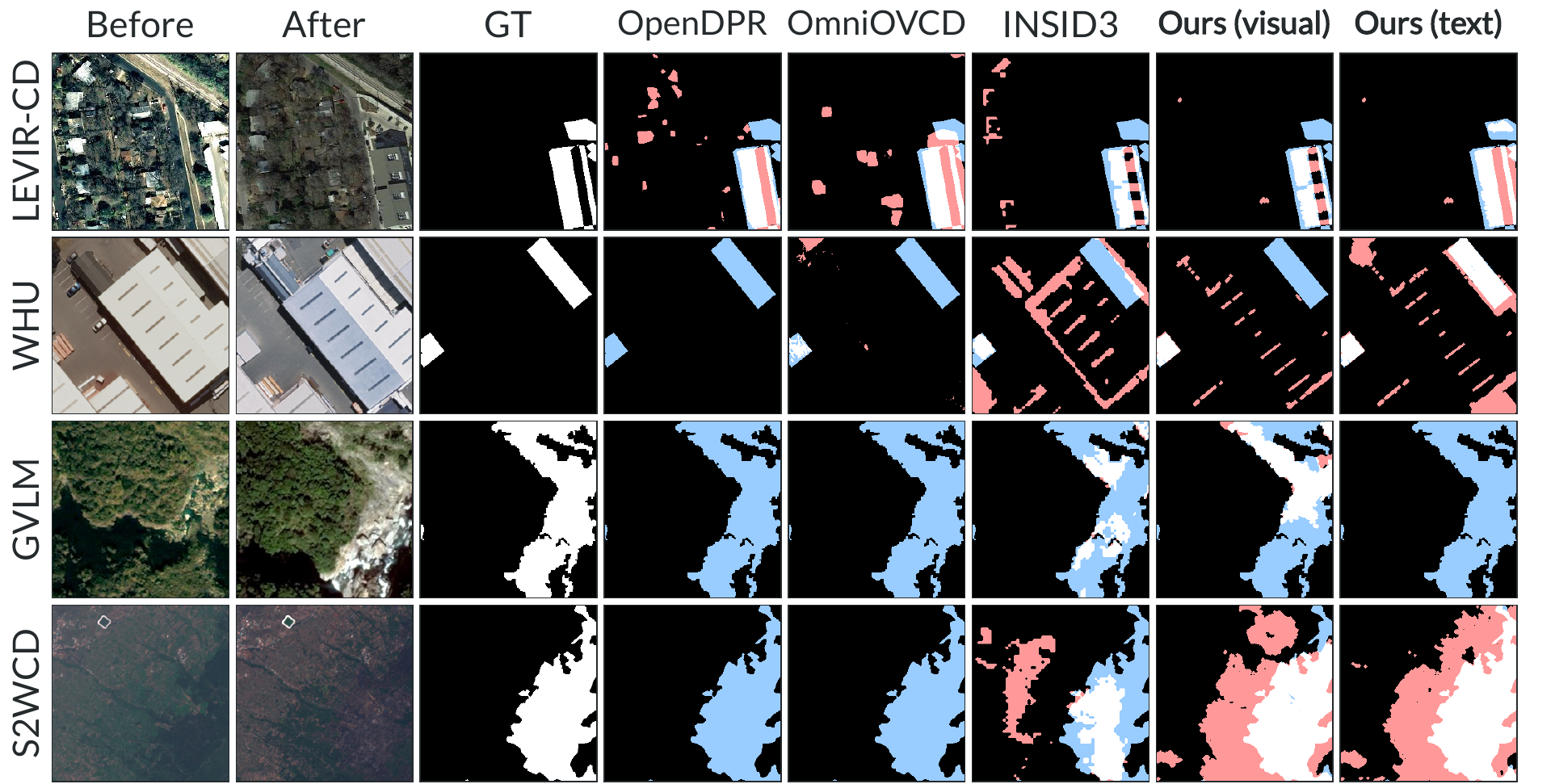}
    \caption{Failure case visualisation. OVTSID3 visualisation was obtained using a text prompt and a visual prompt (the same example as in INSID3). \textcolor{falsePos}{False positives} are red and \textcolor{falseNeg}{false negatives} blue.}
    \label{fig:fail}
\end{figure}

Failure cases of our method are illustrated in \Cref{fig:fail}. We can see that, in the case of buildings, it can exhibit various artefacts if the objects are not completely coherent, \eg, the roof has elements that do not align well.
Another failure point is illustrated by natural disasters, where, in the case of GVLM, the method detects only part of a landslide when prompted with a visual sample, not the full extent, due to large visual deviations from the reference. In the case of S2WCD, we can see that the reference can sometimes have the opposite effect, oversegmenting even regions close to the burn that are not actually affected.

\section{Extended implementation details}
\label{a:impl}

This section provides implementation details for our method and related methods.

\subsection{Our method}

In our experiments, we use pretrained weights from Meta's official links, including both visual and text encoder weights. The ViT-L is pretrained with DINOv3~\cite{simeoni2025dinov3}, and its text encoder is the following: \textcolor{weights}{facebookresearch, dinov3: dinov3\_vitl16\_dinotxt}. The main information is listed in the main paper. Here we list some additional details. 

\paragraph{Core details} 

We use the final layer, as per the official notebook in the above repo, to get the features. Features are normalised before any operation using the L2 norm. We adopted Otsu thresholding for its potential for generalisation and used it for the results in the main paper, but found that a fixed threshold of 0.5 yields comparable performance, suggesting that the choice of change-map threshold is not critical to the results. Images are of the official shape, but are resized to $1024\times1024$ following INSID3~\cite{cuttano2026insid3} and OmniOVCD~\cite{zhang2026omniovcd} for fair comparisons. The metrics are computed at native resolution for all of them.

The debiasing follows the exact same procedure as in INSID3~\cite{cuttano2026insid3}, and the parameters are also the same as there and as stated in the main paper: $r=500$ (also validated there). 

As explained in the main paper, we use agglomerative clustering~\cite{mullner2011modern} with average linkage and $\tau=0.8$ (verified in the main paper).

We use the same CRF~\cite{krahenbuhl2011crf} as INSID3~\cite{cuttano2026insid3} and also apply it to the reported INSID3 results for fair comparison. The gain is modest for both methods (1 point for INSID3, 1.8 for ours), but we keep it following INSID3. Before that, we use bilinear interpolation to resize the mask to match the input image's size.

\paragraph{Prompts}
In the textual path, we use the prompt provided in \Cref{tab:prompts} using the same template as dino.txt~\cite{jose2025dinotxt}. We use our prompts on all other text-based OV methods, as they do not have native config for those (GVLM, S2WCD). We note that our method and most related methods operate on background and foreground prompts, rarely on foreground-only prompts.

We also evaluated our prompts for OmniOVCD~\cite{zhang2026omniovcd}, where it's easiest to perform this swap. It achieves 23.7 IoU and 37.4 F1, compared to 27.1 IoU and 41.8 F1 with its original prompts. This indicates that it works better with its default, so we use that in the main paper for SCD. We did get improvement on building BCD with our prompts, so the results reported in the main paper use the same prompts as our method. The difference may reflect distinct text-alignment mechanisms. Related work typically uses the Segment Anything Model (SAM), which was trained differently from the dino.txt OVTSID3 uses. SAM was designed to segment specific things based on the prompt, while our backbone was trained in a CLIP-like fashion. Based on the results, we argue that our approach is better aligned for use cases such as landslides, where it substantially outperforms SAM-based methods.

Overall, it is beneficial to consider how the object looks from above when describing it in text. And based on the above analysis, and similarly to~\cite{osep2024better}, when using dino.txt or other CLIP-like models, it is better to describe the object with multiple synonyms, especially with our mechanism for taking the target--background margin. We also observed that the method is mostly stable with respect to background prompts, but as additional prompts incur only minimal overhead, it is beneficial to list multiple.

\begin{table*}[t]
    \centering
        \caption{Text prompts. Binary: the concept score is the maximum cosine similarity over the foreground prompts minus the maximum over the background prompts. SECOND: a class score is the mean of the top 3 cosine similarities over its prompts. We use these prompts for our method as well as in BCD for related OVCD text-based methods.}
    \label{tab:prompts}
    \small
    \begin{tabularx}{\linewidth}{@{}lX@{}}
    \toprule
    & Prompts \\
    \midrule
    \multicolumn{2}{l}{\textit{Binary change detection, foreground}} \\
    LEVIR-CD, WHU & house, building, roof, rooftop, flat roof, pitched roof, tiled roof, metal roof \\
    GVLM & dirt, soil, ground, landslide \\
    S2WCD & burn, black burn, burn scar \\
    \midrule
    \multicolumn{2}{l}{\textit{Binary change detection, background (all datasets)}} \\
    All & grass, road, driveway, trees, tree, yard, forest, pavement, field, parking lot, lake, water, river, pond, pool, street, bush, shadow, container, wood, tree canopy, tree crown, treetop, leaves, car, vehicle, truck \\
    \midrule
    \multicolumn{2}{l}{\textit{Semantic change detection (SECOND), one prompt set per class}} \\
    Low vegetation & grass, lawn, meadow, grassland, pasture, green grass, weeds, farmland, cropland, agricultural field, grass field, green field, crop field, grassy area, vegetated land \\
    Non-vegetated ground surface & ground, bare soil, bare ground, soil, dirt, sand, gravel, road, street, asphalt, pavement, concrete, construction site, bare land, dirt road, empty lot, paved area, parking lot, cleared land \\
    Tree & trees, tree, tree canopy, treetop, forest, woodland, tree cover, street trees, tree crowns, dense trees, row of trees, wooded area, forest canopy, cluster of trees \\
    Water & water, river, lake, pond, pool, swimming pool, sea, canal, water surface, reservoir, fish pond, river channel \\
    Building & house, building, roof, rooftop, flat roof, pitched roof, tiled roof, metal roof, house roof, building roof, residential buildings, industrial building, warehouse roof \\
    Playground & playground, running track, athletics track, tennis court, basketball court, football pitch, stadium, sports court \\
    \bottomrule
    \end{tabularx}

\end{table*}

\begin{figure}[!h]
    \centering
    \includegraphics[width=1\linewidth]{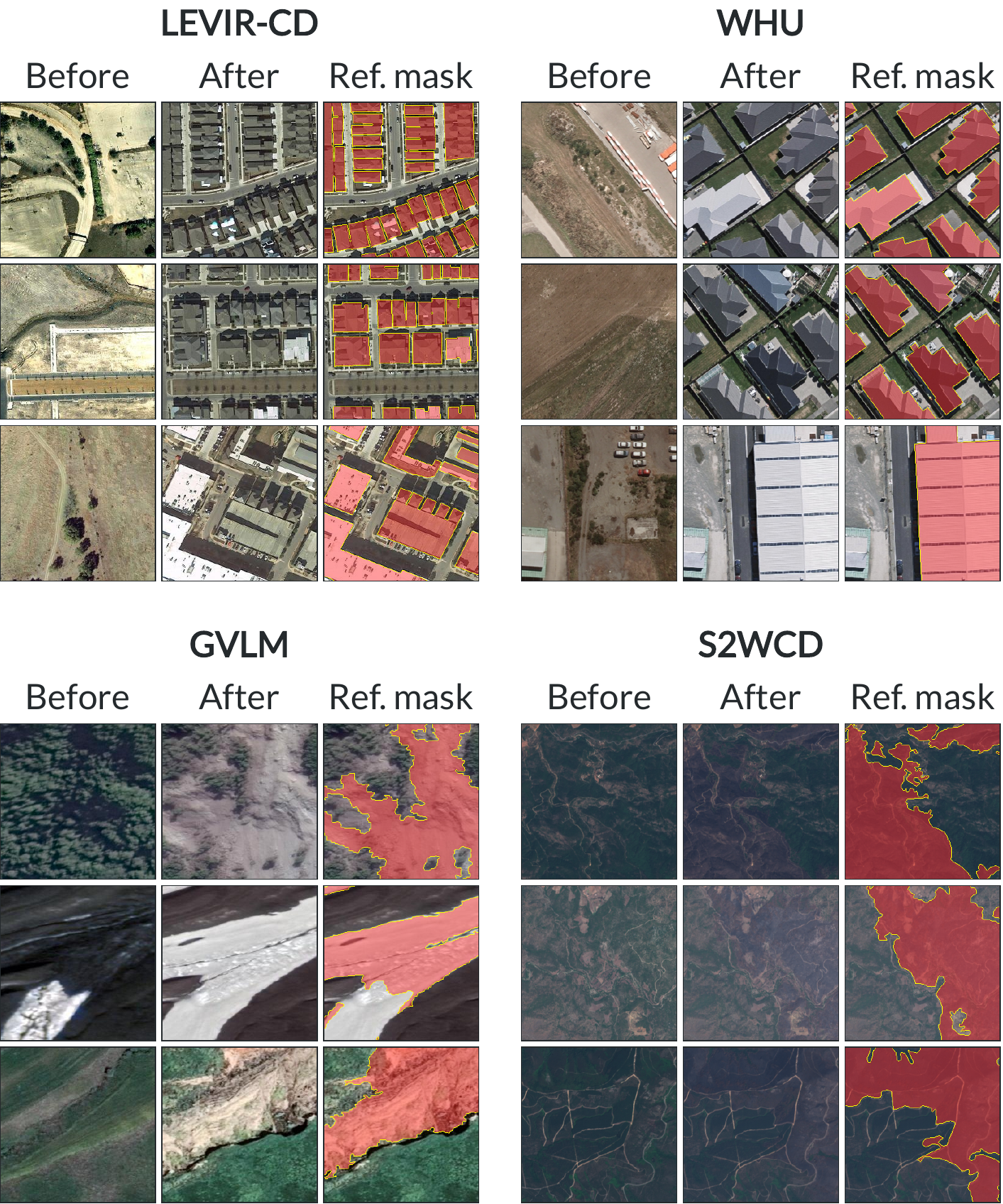}
    \caption{Reference samples (3 random picks for each dataset) used for our method and INSID3.}
    \label{fig:vis_prompt}
\end{figure}

The visual prompt randomly selected for our visual path and INSID3 are presented in \Cref{fig:vis_prompt}. The reference pairs are drawn at random (one per seed) from the training pairs in which 40–60\% of the pixels changed. This simulates a user who provides an example in which the concept covers about half of the image.  Each dataset offers enough such pairs to sample from, and INSID3 and the AnyChange point query receive exactly the same reference pairs.

As stated in the limitations, on some datasets the method is sensitive to the representativeness of the visual prompt. Based on the standard deviation of the main results in \Cref{tab:supp_sota_ovcd}, we can see that for the intended purpose of natural disasters, it is more stable than on buildings, where text is better. This comes from the fact that during pretraining, text is explicitly trained to align with multiple samples referring to the same concepts, while visual samples are obtained from a single image. 

\paragraph{Hyperparameter selection}

We set $\tau = 0.8$, $r=500$, $\delta = 0.01$ for text prompts, while $\delta$ is automatically selected for visual prompts as per the main paper.
Experiments for determining hyperparameters are presented above in \Cref{supp:val}.

\subsection{Related methods}

We use the official code for all related methods in our work where applicable. We also use their default hyperparameters unless stated otherwise below. For SECOND, as stated in the main paper, the results are from~\cite{guo2026opendpr} except for OmniOVCD, which is from the paper itself~\cite{zhang2026omniovcd}. We reran the best method, OpenDPR and reproduced the results. BCD results in Table 1 of the main paper are obtained using the official code for all methods. 

For DynamicEarth M-C-I we use the: SAM~\cite{kirillov2023sam}, DINOv3~\cite{simeoni2025dinov3}, and SegEarthOV~\cite{li2025segearth} and for I-M-C we use: MM-Grounding-DINO~\cite{liu2024grounding}, SAM2~\cite{ravi2025sam2}, and DINO~\cite{caron2021dino} (the codes default). The MCI uses DINOv3, which makes for a fairer comparison. For categories that the original authors never validated (landslides and wildfires), we use the prompt our method uses in every case. OmniOVCD~\cite{zhang2026omniovcd} uses only SAM3. While it does support providing visual exemplars, these have two shortcomings. First, they are in the form of bounding boxes, which are not best suited for concepts like landslides in our case. Second, the exemplar must come from the target image itself, which defeats the purpose of an automated system.

The results for S2WCD on S2C collapse with all seeds, whereas GVLM works fine, so we report that result as is. For AnyChange point query, the point was obtained from the same reference that our visual and INSID3 receive.

We evaluate INSID3~\cite{cuttano2026insid3} using its official implementation and default hyperparameters, without dataset-specific tuning. For the INSID3 + change map baseline, we intersect the union of its two predicted concept masks with the same class-agnostic change mask used by OVTSID3.

\section{Efficiency protocol}
\label{a:eff}

3 different computational efficiency metrics are reported: parameter count, inference time, also expressed as frames per second (FPS), and peak GPU memory usage. Memory is measured using the official PyTorch profiler. 

For inference time, the input is a pair of $1024 \times 1024$ RGB images randomly picked from LEVIR~\cite{chen2020levirStanet} in the case of textual methods and one randomly picked from GVLM~\cite{zhang2023gvlm} for visual methods. This is to simulate the environment where they will most likely be used. The text prompts are the same as our model's in all applicable cases. We do not include the time needed for one-off text (or visual prompt) encoding for any method. We perform 1000 warm-up forward passes, followed by 1000 timed forward passes. This procedure is repeated five times, and the average runtime and memory per forward pass are reported. All measurements are conducted on an NVIDIA A100 GPU.

\subsection{Full efficiency results}

Full results are included in \Cref{tab:perf}. Our method is the second-fastest when text-prompted, after DynamicEarth I-M-C. In terms of parameters, our method has 26M more than the 2nd-best OVCD method, but it is faster. Most of these parameters come from the dino.txt~\cite{jose2025dinotxt} text encoder. The text encoder is not needed in the case of visual prompting, bringing the total down to 303, the same as INSID3, while the inference time of this path is also comparable to it. The longer inference time for the visual path stems from using a random GVLM image instead of LEVIR in that case, which better reflects the situation where one would opt for a visual prompt. Removing clustering, which is done on the CPU, in future work would make the inference time less dependent on input and faster.

\begin{table}[!h]
    \centering
        \caption{Inference cost on one NVIDIA A100 for one $1024\times1024$ image pair (batch 1, time: mean $\pm$ std of repeated runs, memory std below 0.05\, GB, one-off text encoding and reference preparation excluded), with the F1 averaged over the four binary datasets. Parameters include all frozen models (\eg text encoders). OmniOVCD and DynamicEarth are timed with the same prompt list as ours. Our clustering runs on the CPU.}
    \label{tab:perf}
    \setlength{\tabcolsep}{3pt}
    \resizebox{\linewidth}{!}{
    \begin{tabular}{lccccc}
    \toprule
    Method & Time (ms) & FPS & Params (M) & Mem. (GB) & F1 \\
    \midrule
    SCM~\cite{tan2024scm} & 817 $\pm$ 17 & 1.22 & 224 & 0.8 & 24.7 \\
    DynEarth M-C-I~\cite{li2025dynearth} & 2855 $\pm$ 17 & 0.35 & 877 & 7.3 & 40.4 \\
    DynEarth I-M-C~\cite{li2025dynearth} & 418 $\pm$ 4 & 2.39 & 653 & 5.0 & 13.2 \\
    OpenDPR~\cite{guo2026opendpr} & 2333 $\pm$ 1 & 0.43 & 814 & 7.4 & 46.9 \\
    OmniOVCD~\cite{zhang2026omniovcd} & 3359 $\pm$ 2 & 0.30 & 841 & 5.3 & 61.1 \\
    INSID3~\cite{cuttano2026insid3} & 1191 $\pm$ 31 & 0.84 & 303 & 2.3 & 43.2 \\
    \textbf{OVTSID3} (visual) & 1374 $\pm$ 26 & 0.73 & 303 & 2.2 & 65.6 \\
    \textbf{OVTSID3} (text) & 779 $\pm$ 2 & 1.28 & 867 & 2.2 & 70.0 \\
    \bottomrule
    \end{tabular}
    }

\end{table}

\end{document}